\documentclass{article}
\PassOptionsToPackage{numbers}{natbib}

\newcommand{\method}{HABIT}
\usepackage{enumitem}

 \usepackage[eandd, final]{neurips_2026}

\usepackage[utf8]{inputenc} 
\usepackage[T1]{fontenc}    
\usepackage{url}            
\usepackage{booktabs}       
\usepackage{amsfonts}       
\usepackage{nicefrac}       
\usepackage{microtype}      
\usepackage{xcolor}         

\usepackage{amsmath}

\usepackage{float}

\usepackage{graphicx}
\usepackage{subcaption}
\usepackage{caption}
\definecolor{myblue}{RGB}{30,120,255}

\usepackage[
    colorlinks=true,
    urlcolor=myblue,
    linkcolor=black,
    citecolor=black
]{hyperref}

\title{HABIT: Human-Aware Behavior and Interaction Training Dataset for Robot Manipulation}

\author{
  \vspace{0.4em}
  \begin{tabular}[t]{ccc}
    \makebox[3.9cm]{Jaehwi Song} & \makebox[3.9cm]{Suchae Jeong} & \makebox[3.9cm]{Byeongguk Jeon} \\
    \mdseries Config, KAIST & \mdseries Config, KAIST & \mdseries Config, KAIST \\[1.4em]
    Sungdong Kim & Hyungmok Son & Kimin Lee \\
    \mdseries Config, KAIST & \mdseries Config & \mdseries Config, KAIST
  \end{tabular}
  \AND
  \vspace{-2.1em}\\
  {\url{https://habit-dataset.github.io}}
}

\begin{document}

\maketitle
\vspace{-1.3em}
\begin{abstract}
Large-scale demonstration datasets have been central to recent progress in general-purpose robot policies. However, existing datasets are collected in human-absent settings, and policies trained on such data may perform tasks competently in isolation but fail to exhibit human-aware behaviors. To address this gap, we introduce HABIT, a large-scale robot demonstration dataset for human-present environments. We organize tasks into three roles capturing distinct modes of human-robot interaction: Collaborator, where human and robot jointly accomplish a task; Coworker, where they pursue separate tasks in a shared space; and Supervisor, where the human directs the robot. The dataset comprises over 10K episodes and over 160 hours across 60 tasks. Our experiments show that training on human-present data elicits human-aware behaviors that robot-only data fails to produce: spatiotemporal synchronization in Collaborator tasks, yielding in Coworker tasks, and gesture grounding in Supervisor tasks. Moreover, training on HABIT enables rapid adaptation to new human-robot interaction tasks. By introducing human presence as a new axis of dataset diversity, HABIT extends robot policies to environments shared with humans.
\end{abstract}
\section{Introduction}
\label{sec:intro}

Data-driven approaches have emerged as a promising direction for training robotic manipulation policies~\citep{ rt1,jang2022bc}. Recent robot datasets have grown in scale and diversity by collecting data across multiple embodiments~\citep{rh20t,open_x_embodiment, wu2024robomind, robonet}, tasks~\citep{agibot_world,droid, walke2023bridgedata}, and even human demonstration videos~\citep{ego4d, ego_exo4d, punamiya2026egoverse, zheng2026egoscale}. Trained on these increasingly large and diverse manipulation datasets, vision-language-action (VLA) models~\citep{groot,pi0, openvla, octo,rt2} and world action models (WAMs)~\citep{cosmospolicy,liang2025video, mimicvideo, dreamzero} generalize across scenes, embodiments, and tasks.

However, these datasets are usually collected in human-absent settings, with the robot acting as the sole agent in the scene~\citep{agibot_world, droid, walke2023bridgedata}. As a result, policies trained on such data are unlikely to perform well in the environments where these robots are meant to be deployed. In homes, factories, and other shared workspaces, a robot must coordinate with co-present humans (e.g., following their cues, anticipating their motions, and avoiding collisions~\citep{bauer2008human,scholtz2003, onnasch2021, pascher2023}). These behaviors are missing from human-absent data not because they are hard to learn, but because they cannot be demonstrated without a human in the scene. For example, a robot cannot learn to hand over a tool if no one is there to receive it, nor to pause for a reaching hand if no hand ever reaches. This gap motivates dedicated datasets that explicitly capture human–robot interaction dynamics and encode collaborative, human-aware behaviors.

In this work, we introduce \textbf{\method} (Human-Aware Behavior and Interaction Training dataset), a large-scale robot demonstration dataset explicitly designed for human-present environments. In every episode of \method, a co-present human shares the workspace with the robot. The dataset comprises 10{,}563 episodes and 164 hours of bimanual manipulation, spanning 60 tasks. Tasks are organized along three interaction roles that capture distinct dependencies between human and robot: \textit{Collaborator}, where human and robot jointly accomplish a shared task; \textit{Coworker}, where human and robot pursue separate tasks within a shared space; and \textit{Supervisor}, where the human observes and directs the robot. To elicit specific human-aware behaviors such as yielding and gesture-following, we carefully design our collection protocols, while varying other conditions to support generalization.

We verify the effectiveness of \method~by fine-tuning two open-source VLAs, $\pi_{0.5}$ \citep{pi05} and GR00T N1.6 \citep{groot}, on a representative six-task subset, and comparing against a matched Robot-only baseline collected without a co-present human. \method~improves task success rates for both models, with the largest gains on tasks where role-specific coordination is most critical. More notably, training on \method~gives rise to human-friendly behaviors that emerge directly from data: proactive yielding and collision avoidance under the {\em Coworker} role, gesture grounding under {\em Supervisor}, and spatiotemporal synchronization under {\em Collaborator}. These behaviors reflect the model's internalization of social context when trained on human-present demonstrations. 
Finally, we show that $\pi_{0.5}$ trained on \method~adapts rapidly to new human-robot interaction tasks.
We believe \method~as a stepping stone toward robot foundation models that are not merely capable, but genuinely safe and socially compatible in the human-inhabited environments where they will ultimately be deployed.

\begin{figure}[t]
  \centering
  \includegraphics[width=1.0\linewidth]{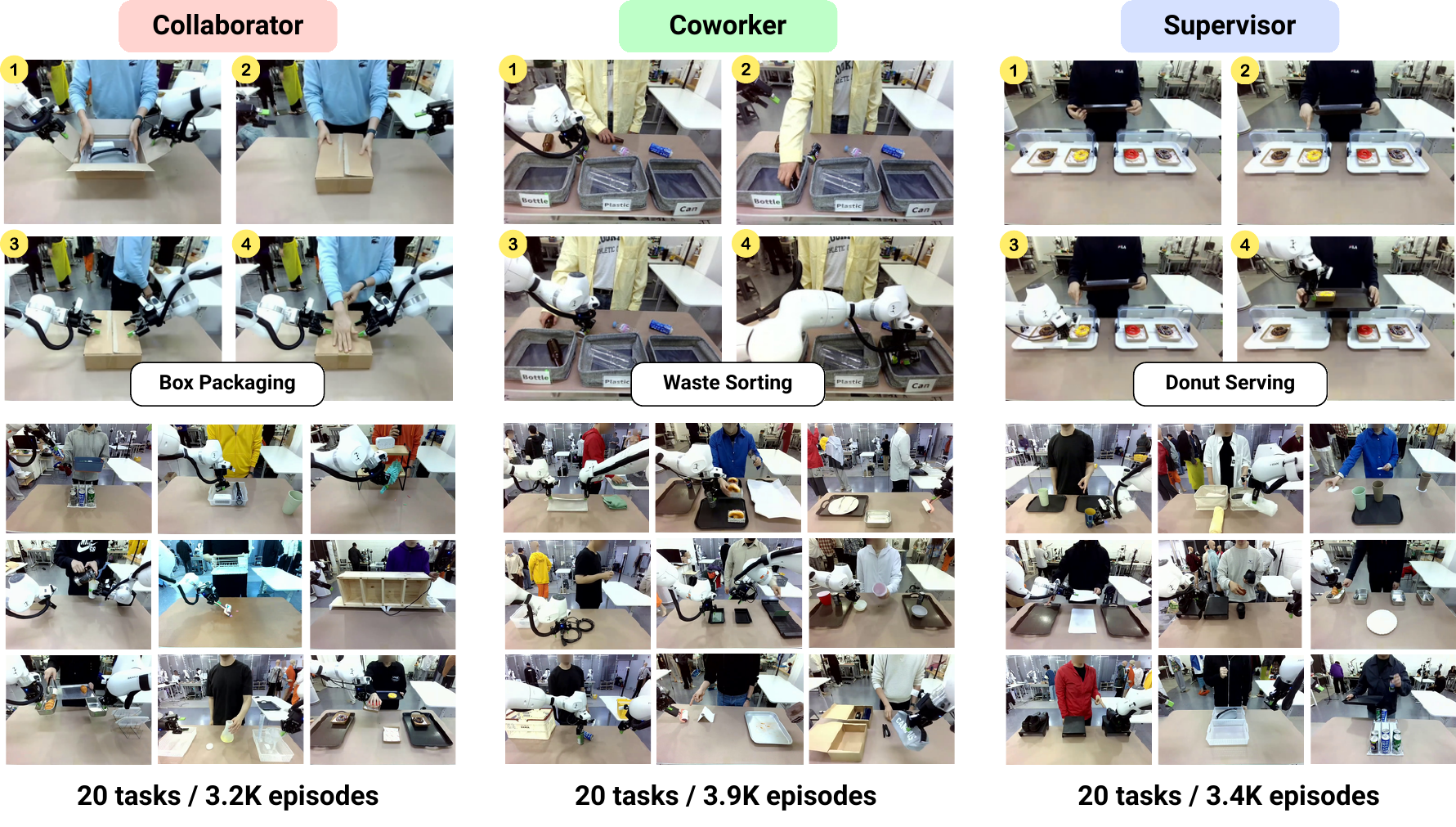}
  \caption{\method~comprises 164 hours of human-robot interaction demonstrations across 60 tasks spanning three roles (\emph{Collaborator}, \emph{Coworker}, and \emph{Supervisor}) defined by how the human and robot interact within a subtask.}
\label{fig:dataset_overview}
\vspace{-0.1in}
\end{figure}

\begin{figure}[t]
  \centering
  \includegraphics[width=1.0\linewidth]{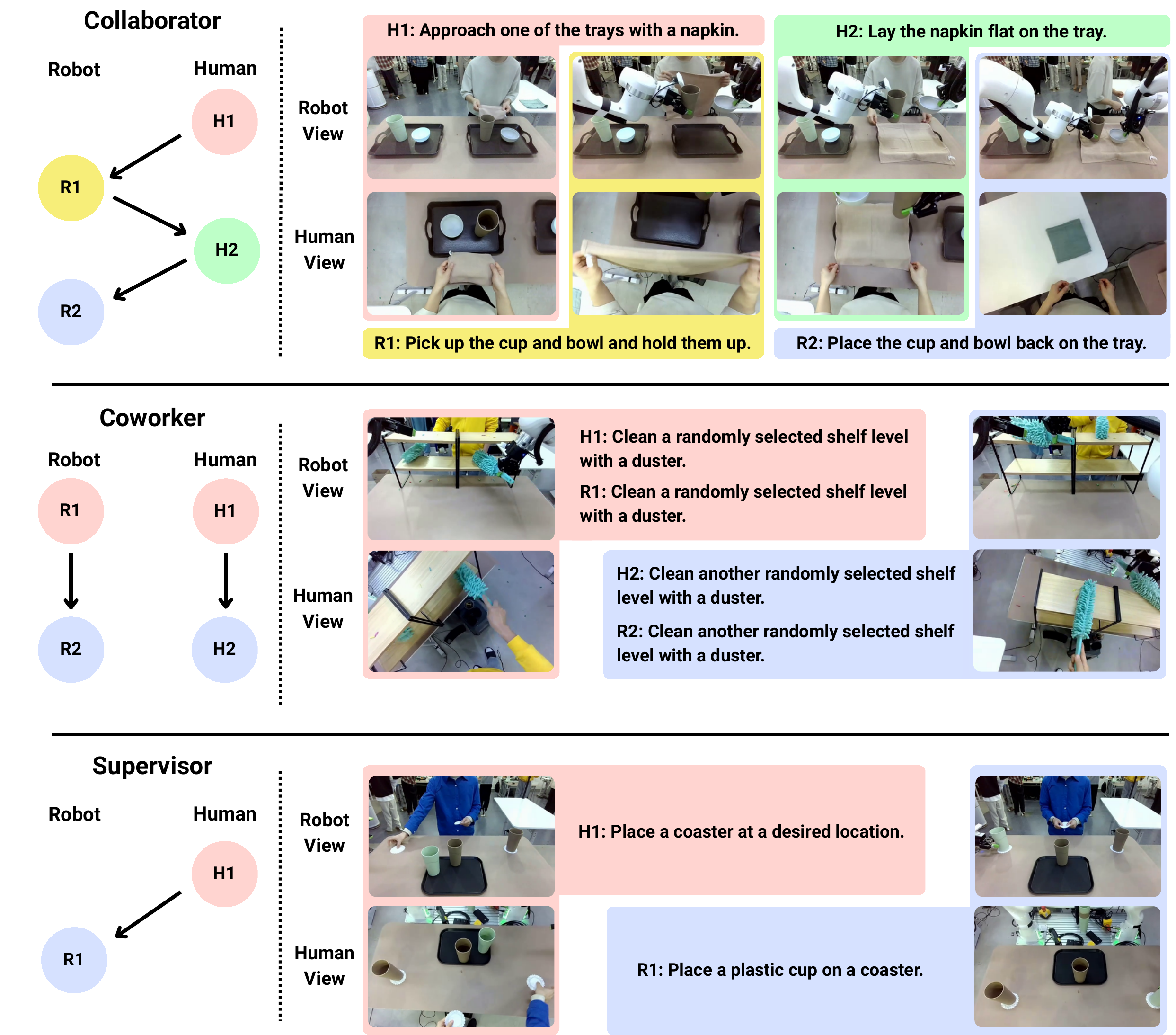}
  \caption{Representative examples of task workflows with their subtask sequences. For each row, the workflow is shown on the left, with the robot and human views at the corresponding stages of execution on the right.}
  \label{fig:task-structure}
  \vspace{-0.1in}
\end{figure}

\begin{figure}[t]
    \centering

    \begin{subfigure}[b]{0.59\linewidth}
        \centering
        \includegraphics[width=\linewidth]{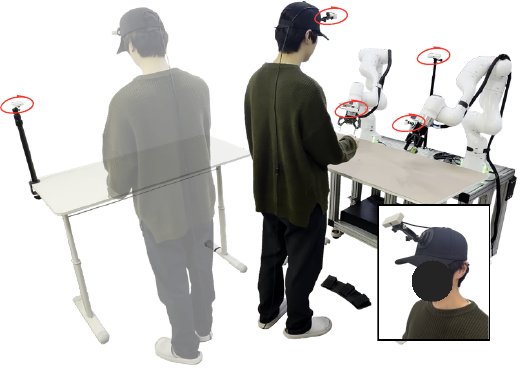}
        \caption{Workspace setup}
        \label{fig:hardware}
    \end{subfigure}
    \hfill
    \begin{subfigure}[b]{0.38\linewidth}
        \centering
        \includegraphics[width=\linewidth]{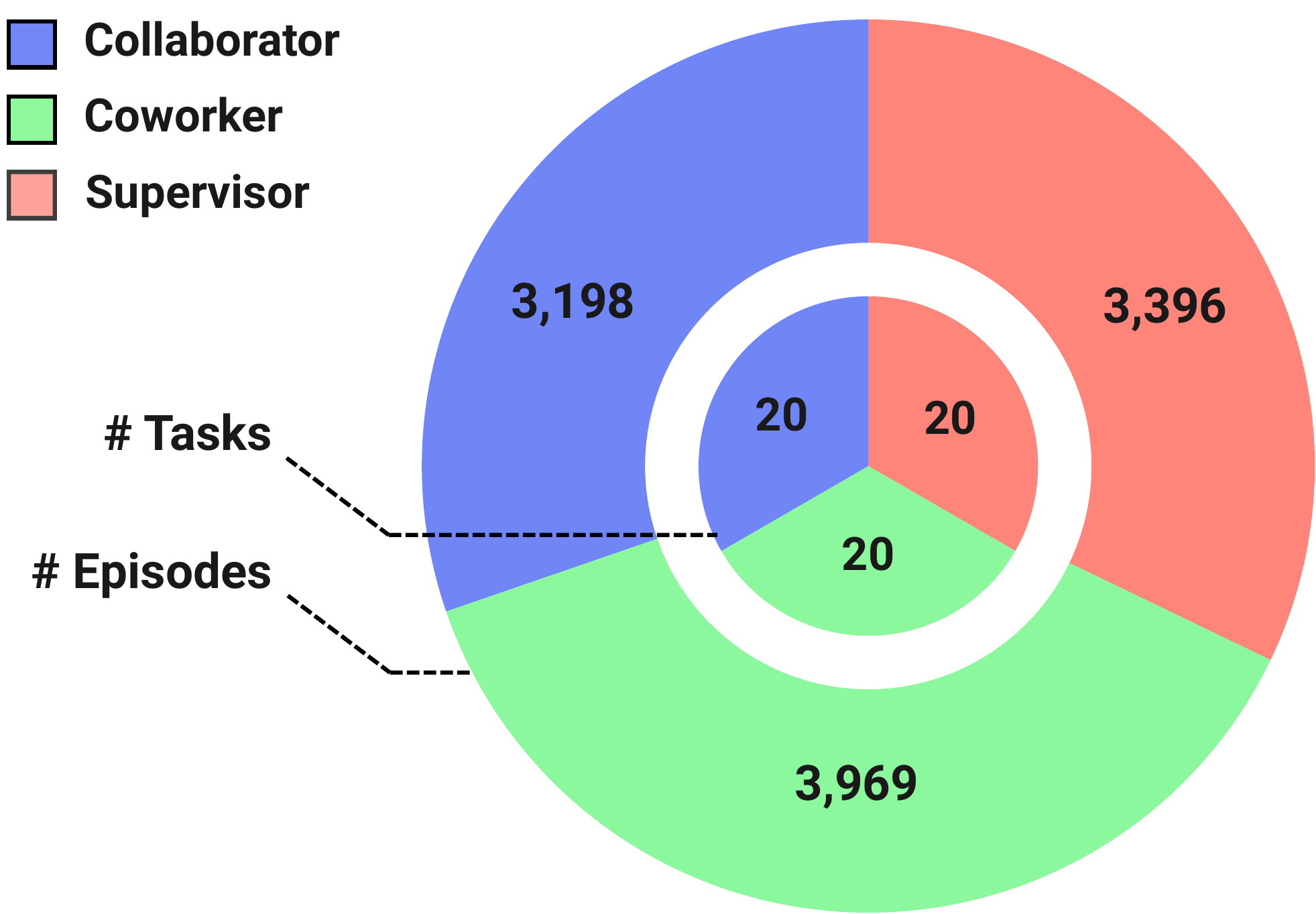}
        \vspace{0.6cm}
        \caption{Per-category distribution}
        \label{fig:dataset_stats}
    \end{subfigure}

    \caption{
    (a) The collection unit includes both the human and the robot agent, along with five RGB cameras marked in red. We use three cameras for robot manipulation and add one ego-view and one exo-view camera to capture the full task progression of both the human and the robot.
    (b) Distribution of tasks and episodes across human-role categories. Each category contains 20 tasks, with an approximately balanced number of episodes.}
    \label{fig:setup}
\end{figure}

\section{\method~Dataset}
\label{sec:dataset}

In this section, we introduce \method, a large-scale robot demonstration dataset for environments shared with humans. Section~\ref{sec:task-desgin} describes the task design, Section~\ref{sec:hardware} shows the hardware setup, Section~\ref{sec:data-collection} details the data collection protocol, and Section~\ref{sec:stats} reports dataset statistics.

\subsection{Task Design}
\label{sec:task-desgin}

Our tasks are designed to capture diverse scenarios of human–robot interaction in co-present settings, where a human and a robot share the same workspace.\footnote{For simplicity, we focus on one-robot, one-human environments. Extending our framework to multi-robot or multi-human settings is an interesting direction for future work.} Following prior work~\citep{scholtz2003, onnasch2021, pascher2023}, we consider three categories of human roles: Collaborator, Coworker, and Supervisor. In {\em Collaborator} tasks, the human and robot jointly accomplish a shared goal through direct physical interaction (e.g., handing over an object or jointly holding a bucket), and the robot must coordinate with the human both spatially and temporally. {\em Coworker} tasks likewise involve a shared goal and workspace, but without direct physical contact; here, the robot must avoid collisions with the human to ensure safety. Finally, in {\em Supervisor} tasks, the human directs the robot through explicit cues such as gestures or demonstrated actions, and the robot must infer the human's intent from visual input alone. Representative examples of each category are illustrated in Figure~\ref{fig:dataset_overview}.

Early in development, we found that providing only a task instruction (e.g., ``clean the shelf with a duster together'') was too ambiguous to elicit consistent demonstrations: episodes under the same instruction varied substantially in how the human and robot divided and sequenced subtasks. To address this, we introduce the {\em task workflow}, which specifies how the human and robot should interact at the subtask level. Formally, a task workflow is a directed graph in which nodes correspond to subtasks performed by the human or the robot, and edges encode the order in which these subtasks must be executed. Each node is labeled $H_i$ or $R_i$, where the letter denotes the agent (human or robot) and  $i$ indexes the subtask within that agent's sequence. We instruct both human and robot operators to execute each task by following its workflow, enforcing consistency across episodes. Figure~\ref{fig:task-structure} illustrates an example workflow for each role. Note that workflows for {\em Coworker} tasks are dominated by single-agent edges, reflecting the structure of the role itself: two independent subtask chains that share only the workspace.

\subsection{Workspace Setup}
\label{sec:hardware}

Figure~\ref{fig:setup} shows our workspace setup. We use two Franka Research 3 arms with Robotiq grippers, teleoperated via the controllers of a Meta Quest 3 headset. To establish a shared workspace with a human, we set up two tables: a {\em front table} placed between the human and the robot, and a {\em side table} located beside the human. The front table serves as the shared workspace, while the side table is reserved for human-only activities.

We record image observations with five synchronized cameras. Three are mounted on the robot side: one on each wrist, and a third providing an egocentric view angled forward to capture both the human and the shared workspace. The remaining two cameras are dedicated to the human-side area. The first is mounted on the human's head and captures their egocentric perspective, conveying intentions such as pointing gestures whose precise referents are often difficult to resolve from robot-side views alone. The second is positioned to observe the entire human–robot workspace, providing a holistic view of the interaction.\footnote{For all experiments, we use only the robot-side cameras to train robot policies, for computational efficiency. We include all five views in the released data to support future research.}

Further details on teleoperation and camera specifications are provided in Appendix~\ref{app:hardware}.

\subsection{Data Collection Protocol} \label{sec:data-collection}

To effectively capture human-aware behaviors in the collected data, our protocol goes beyond merely placing a human in the workspace alongside the robot. We structure the collection process to elicit specific human-aware behaviors, and vary other conditions to support generalization.

\paragraph{Reactive interaction.} The task workflow can introduce a failure mode: because operators know the subtask sequence in advance, they can pre-execute their next subtask from memory rather than in response to their partner. For example, the robot operator may begin moving the arm toward the target object before the human operator points to it. This is problematic because the cue that triggers the robot's action then falls outside the camera input, making the behavior unlearnable. To address this, we adopt reactive interaction as a core principle of data collection: each operator acts only after directly observing the partner's behavior, and we prohibit any coordination signal that is not captured in the recorded observations. Every demonstrated action is therefore grounded in cues that are also available to the policy, making the resulting demonstrations learnable.

\paragraph{Behavior elicitation.}We adopt three design choices to elicit specific human-aware behaviors that reactivity alone does not guarantee. 
\begin{itemize} [leftmargin=1.5em, itemsep=1mm, topsep=1pt]
    \item \textit{Yielding under safety-first priority.} We treat human safety as the overriding constraint during data collection. Whenever the robot is about to collide with the human or with human-held objects, the robot operator decisively retracts the arms rather than continuing the trajectory, so that yielding is recorded as the robot's default response. 
    \item \textit{Temporal adaptation.} The human operator's movement speed is deliberately varied across episodes, so that policies trained on the data must align their tempo with the partner rather than execute at a fixed cadence.
    \item  \textit{Gesture grounding.} For tasks where the human directs the robot through gestures, the human operator samples the wait time before pointing from pre-defined bins (see Appendix~\ref{app:task-details} for details), forcing policies to attend to the gesture itself rather than acting prematurely on the language instruction alone.
\end{itemize}

\paragraph{Additional diversification.}Beyond the elicitation choices above, we vary collection conditions to prevent overfitting to incidental factors. Within each task, we vary clothing color across episodes and randomize the order in which objects are manipulated. Across tasks, the dataset spans multiple human operators with different body types. Together, these variations support the out-of-distribution (OOD) evaluation in Appendix~\ref{app:ood}.

\begin{figure}[t]
    \centering
    \newlength{\panelsep}\setlength{\panelsep}{0.012\linewidth}
    \newlength{\panelw}\setlength{\panelw}{\dimexpr(\linewidth-2\panelsep)/3\relax}
    \begin{subfigure}[b]{\panelw}
        \centering
        \includegraphics[width=\linewidth]{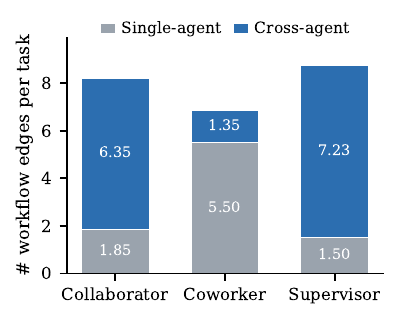}
        \caption{Interaction structure by role}
        \label{fig:interaction}
    \end{subfigure}%
    \hspace{\panelsep}%
    \begin{subfigure}[b]{\panelw}
        \centering
        \includegraphics[width=\linewidth]{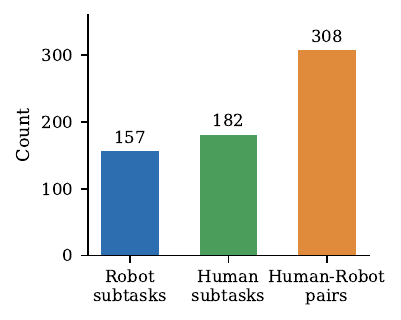}
        \caption{Subtask diversity}
        \label{fig:diversity}
    \end{subfigure}%
    \hspace{\panelsep}%
    \begin{subfigure}[b]{\panelw}
        \centering
        \includegraphics[width=\linewidth]{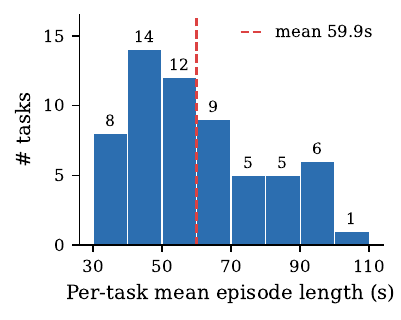}
        \caption{Task duration distribution}
        \label{fig:duration}
    \end{subfigure}
    \caption{
    Dataset statistics.
    (a) Per-role workflow composition, with single-agent and cross-agent ordering edges averaged over the tasks in each role.
    (b) Subtask diversity,  where 157 distinct robot subtasks and 182 distinct human subtasks combine into 308 unique human-robot subtask pairs.
    (c) Distribution of per-task mean episode length over the 60 tasks.}
    \label{fig:datastats}
\end{figure}

\subsection{Dataset statistics}
\label{sec:stats}
\method~contains 10{,}563 episodes and 164.19 hours of bimanual manipulation across 60 tasks, averaging 3.67 robot and 4.30 human subtasks per episode. To quantify how tightly the two agents are coupled, we classify every ordering edge in a task workflow as single-agent when it connects two subtasks of the same agent and cross-agent when it connects a human subtask to a robot subtask. The majority of edges are cross-agent, so most ordering constraints bind the human and robot together rather than running within one agent's chain. As shown in Figure~\ref{fig:interaction}, this composition varies sharply across roles and recovers the role definitions of Section~\ref{sec:task-desgin}. Collaborator and Supervisor tasks are dominated by cross-agent edges, capturing tight physical coordination and gesture following, whereas Coworker tasks are dominated by single-agent edges, reflecting two largely independent chains that share only the workspace. 

Beyond this structure, Figure~\ref{fig:diversity} shows the subtask diversity of HABIT, which spans a wide variety of human-robot interaction scenarios. Across the dataset, the human performs 182 distinct subtasks and the robot performs 157, and together these form 308 unique human-robot subtask pairs. \method~provides both the human and robot subtask annotations at this granularity, extending the robot-only annotations of prior datasets to the co-present human. Figure~\ref{fig:duration} shows the distribution of per-task mean episode length, which ranges from short single-step interactions to long multi-step interactions. This spread follows from a task design that aims to capture diverse real-world scenarios rather than a single fixed interaction pattern.

\section{Evaluation Framework}
\label{sec:evaluation}

When a human is co-present in the workspace, evaluating the robot requires more than measuring manipulation success. A successful policy must not only manipulate objects correctly, but also coordinate with the human according to the task workflow while maintaining human safety. Our evaluation framework therefore jointly assesses manipulation performance, workflow compliance, and safety under human-robot interaction. To capture different interaction challenges, we evaluate policies using success criteria on role-specific evaluation tasks.

\paragraph{Success criteria.}
A task in \method~is structured as a workflow of human and robot subtasks, as illustrated in Figure~\ref{fig:task-structure}. We score each robot subtask on three binary criteria:
\begin{itemize} [leftmargin=1.5em, itemsep=1mm, topsep=1pt]
    \item {\em Manipulation}: The robot completes the physical manipulation required by the subtask and achieves the intended object state.

    \item {\em Workflow compliance}: The robot satisfies the structural condition imposed by the task workflow, including required ordering, spatiotemporal synchronization, or cue following.

    \item {\em Human safety}: The robot completes the subtask without human--robot collision or contact with human-held objects.
\end{itemize}

We refer to one execution of the policy on a task as a trial, denoted by $\tau$, and define $\tau$ as successful if every robot subtask in $\tau$ satisfies the three criteria above. The success rate over $N$ independent trials per condition is
\begin{equation}
\text{Success rate} = \frac{1}{N}\sum_{i=1}^{N} \mathbf{1}[\tau_i \text{ succeeds}],
\quad
\mathbf{1}[A] =
\begin{cases}
1, & \text{if } A \text{ is true},\\
0, & \text{otherwise}.
\end{cases}
\label{eq:hsr}
\end{equation}

Success rates are reported as percentages throughout Section~\ref{sec:experiments}.

\paragraph{Evaluation tasks.}
For evaluation, we deliberately select a representative subset of 6 tasks from \method, with two tasks per role. This subset is chosen to highlight the key challenge of each role in different task settings. For \emph{Collaborator}, we use tasks that require tight spatial and temporal coordination on a shared activity: Table Serving, where the robot lifts the dishware on whichever of two trays the human approaches so that the human can lay a napkin underneath, and Shelf Cleaning, where the human lifts objects off a tier so that the robot can dust it, and the robot returns the duster after cleaning all tiers. For \emph{Coworker}, we use tasks that vary the amount of workspace overlap during parallel work: Waste Sorting as a moderate-overlap setting, averaging 2 yielding events per trial, and Box Packaging as a high-overlap setting, averaging 3 yielding events per trial. For \emph{Supervisor}, we use tasks that require interpreting human pointing cues in different placement contexts: Donut Serving, where the robot places the indicated donut on a tray, and Food Storage, where the robot places bread in the indicated container. An overview of all evaluation tasks is shown in Figure~\ref{fig:eval_tasks}, and full details of the evaluation set are provided in Appendix~\ref{app:task-details}.

\begin{figure}[H]
  \centering
  \includegraphics[width=\linewidth]{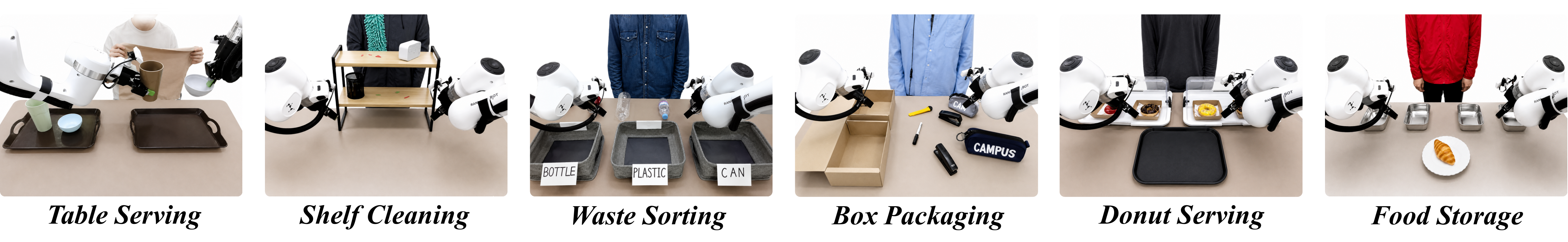}
  \caption{Representative evaluation tasks across the three human roles. Collaborator: \emph{Table Serving} and \emph{Shelf Cleaning}. Coworker: \emph{Waste Sorting} and \emph{Box Packaging}. Supervisor: \emph{Donut Serving} and \emph{Food Storage}.}
  \label{fig:eval_tasks}
  \vspace{-0.1in}
\end{figure}
\section{Experiments}
\label{sec:experiments}

We design our experiments to investigate the following:
\begin{itemize} [leftmargin=1.5em, itemsep=1mm, topsep=1pt]
\item Can our \method~dataset improve task success rates on representative human-robot interaction tasks (Figure~\ref{fig:main-results})?
\item Do robot policies trained on \method~exhibit emergent human-aware behaviors, such as collision avoidance, when we examine their failure cases (Figure~\ref{fig:failure-analysis})?
\item Does mid-training a VLA model on \method~enable more sample-efficient adaptation when fine-tuning with limited downstream data (Figure~\ref{fig:scaling})?
\end{itemize}

\subsection{Experimental Setup}
\label{sec:exp-setup}

\paragraph{VLA models and training.}
We fine-tune two open-source VLAs on \method's bimanual Franka configuration: $\pi_{0.5}$ \citep{pi05} and GR00T N1.6 \citep{groot}. Our goal is to evaluate \method~as a data resource, not to develop a new training method or compare architectures, so within each model we hold training steps (5{,}000) and batch size (128) fixed across Robot-only and \method~fine-tunes, with other hyperparameters tailored to each model. Dataset is the only factor that differs between Robot-only and \method~within each model. Public checkpoints are not available for our bimanual Franka morphology, so zero-shot baselines are not included. Training details are in Appendix~\ref{app:model-training}.

\paragraph{Baselines.}
For each (task, model) pair, we compare \textbf{\method} against \textbf{Robot-only}. \method~consists of our demonstrations collected with a co-present human, while Robot-only is the conventional baseline of teleoperated demonstrations collected without a co-present human. The Robot-only condition uses the same task, environment, and teleoperator as \method, with the only difference being the absence of the human operator.
For Coworker tasks the robot completes its own portion of the parallel work in an empty workspace, for example
clearing only the cans on its side in Waste Sorting or packing only the items it is assigned in Box Packaging.
For Supervisor tasks, both conditions receive the same indexed language instruction (e.g., ``place the bread in the $k$-th container from the left''), with \method~additionally providing a co-located pointing gesture. Including the index in both conditions ensures the comparison isolates the contribution of the human's gesture rather than confounding it with whether the target is specified at all. Robot-only baselines are not applicable for Collaborator tasks since the task itself requires a human partner, so those cells are marked N/A. Training data averages 200 episodes per condition. Full episode counts and hours per task are in Appendix~\ref{app:training-data}.

\paragraph{Evaluation protocol.}

Each (task, model, condition) cell is evaluated over $N=20$ trials by the same human operator, following a predefined task-specific evaluation protocol described in Appendix~\ref{app:eval-protocol}. We additionally evaluate out-of-distribution (OOD) robustness to human-centric distribution shifts (clothing, body shape) in Appendix~\ref{app:ood}. All trials are scored using the success rate defined in Eq.~\ref{eq:hsr}.

\subsection{Main Results}
\label{sec:exp-main}


\begin{figure}[t]
  \centering
  \begin{subfigure}{\linewidth}
    \centering
    \includegraphics[width=0.98\linewidth]{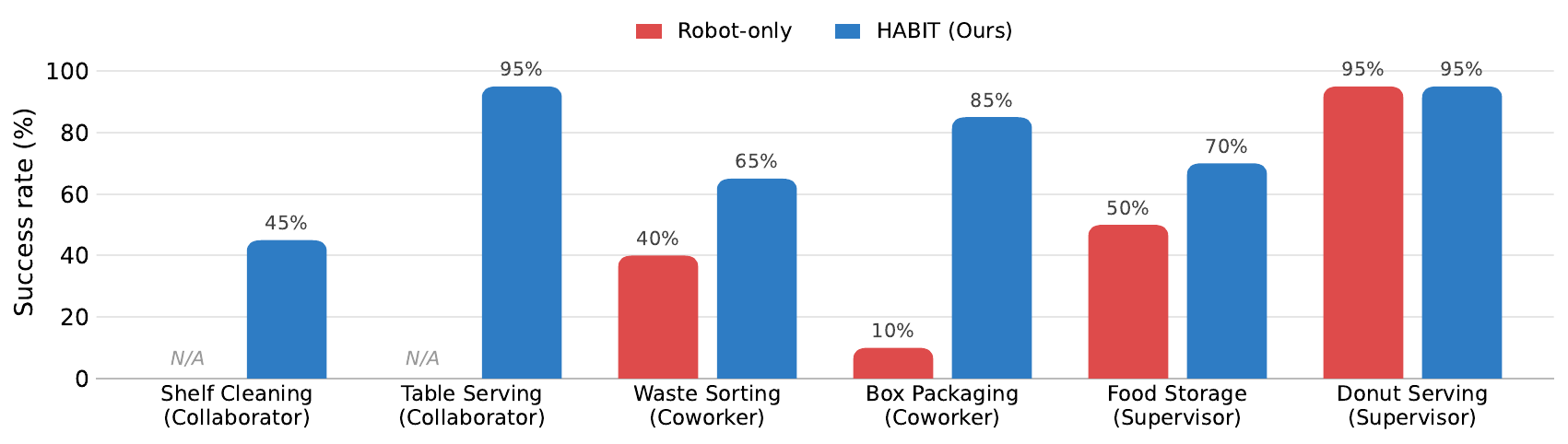}
    \caption{$\pi_{0.5}$}
    \label{fig:main-results-pi05}
  \end{subfigure}
  \begin{subfigure}{\linewidth}
    \centering
    \includegraphics[width=0.98\linewidth]{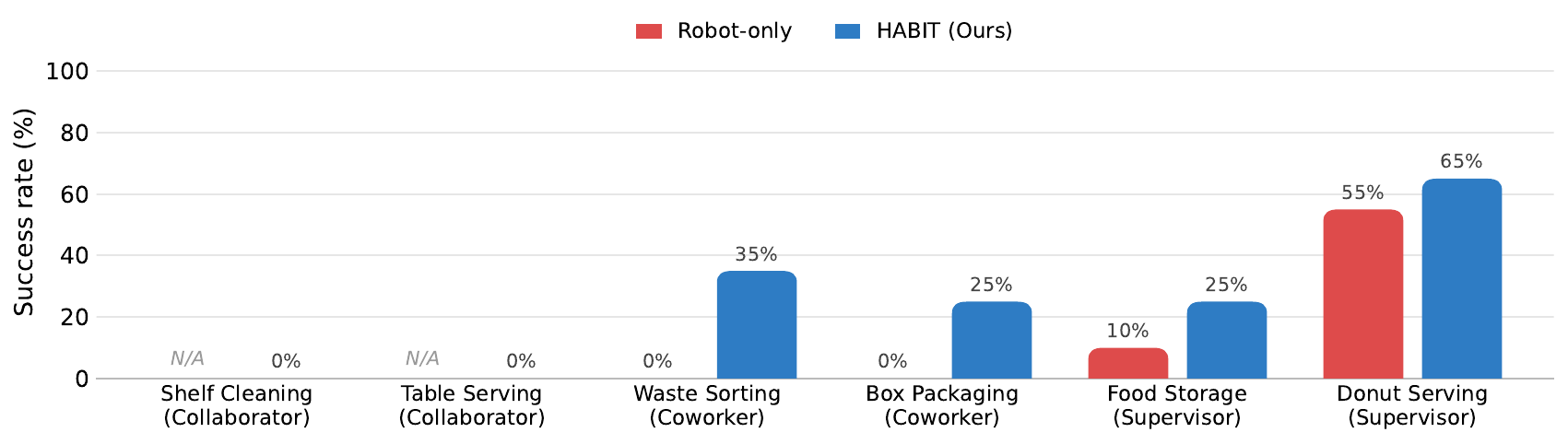}
    \caption{GR00T N1.6}
    \label{fig:main-results-groot}
  \end{subfigure}
  \caption{Success rate across six evaluation tasks for (a) $\pi_{0.5}$ and (b) GR00T N1.6. Each cell reports the mean over 20 trials. The Robot-only condition is not applicable for Collaborator tasks, which require a human partner.}
  \label{fig:main-results}
\end{figure}

Figure~\ref{fig:main-results} reports success rates across the six tasks for $\pi_{0.5}$ and GR00T N1.6, each trained on \method~and on Robot-only. The two Collaborator tasks inherently require a co-present human operator, so Robot-only is not applicable and is omitted from evaluation.

Training $\pi_{0.5}$ on \method~improves success rates over Robot-only on every comparable task, with the largest gains on Coworker tasks, where workspace overlap requires reactive yielding to resolve path conflicts with the human. 
The two Supervisor tasks reveal a more nuanced picture. 
On Donut Serving, \method~and Robot-only perform comparably: the task is a single-stage reach to the target donut, and the indexed language instruction alone is sufficient to disambiguate it. 
Food Storage, by contrast, requires a two-stage trajectory (i.e., the robot first picks up the bread, then proceeds to one of four containers) and the human's pointing gesture provides a co-located visual cue at this branching point, which we attribute as the source of \method’s gain on this task.

The same pattern holds for GR00T N1.6 across the four comparable cells, with Coworker tasks again showing the largest gains. GR00T's absolute success rates are lower than $\pi_{0.5}$'s across all tasks, but the relative benefit of \method~over Robot-only is consistent across both models, suggesting that the gains stem from the dataset rather than model-specific factors. 
These results highlight a gap in standard robot learning pipelines, which rely on data collected in human-absent settings. 
While such data yields capable task performers, the consistent gains from \method~indicate that demonstrations collected with a co-present human carry signal that human-absent data cannot provide.

\subsection{Failure Analysis}
\label{sec:failure}

\begin{figure}[t]
\vspace{-0.1in}
  \centering
  \includegraphics[width=0.95\linewidth]{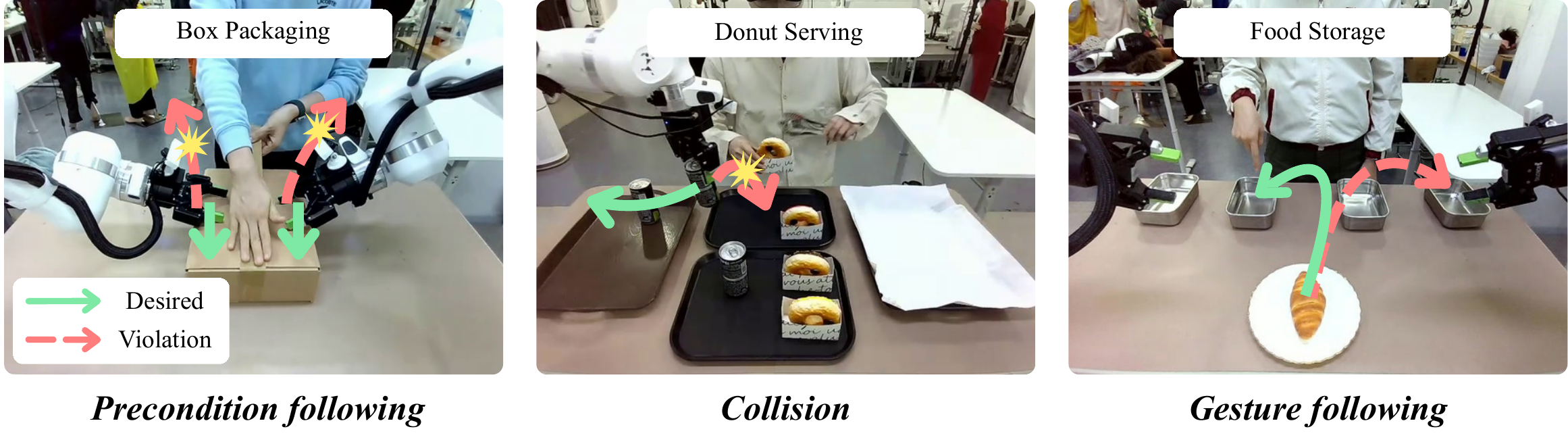}
    \caption{Role-specific failure cases for the Collaborator (left), Coworker (middle), and Supervisor (right) roles. Green arrows denote desired trajectories; red arrows denote violations.}
    \label{fig:failure-case}
    \vspace{-0.1in}
\end{figure}

Because our tasks involve human-robot interaction, failure cases extend beyond simple manipulation errors. In particular, we find that models exhibit a distinct failure mode for each role (Figure~\ref{fig:failure-case}). Precondition violation (i.e., pre-executing the next subtask before the human has completed theirs) is most prevalent in Collaborator tasks, where human and robot subtasks are tightly coupled. Collisions dominate Coworker tasks, where the human and robot act simultaneously in a shared workspace in parallel. Gesture-following failure is most common in Supervisor tasks, where correct robot behavior hinges on grounding the human's cue in the visual scene. We refer to these three role-dependent failures collectively as role-specific failures, distinguishing them from manipulation failures (e.g., grasp slips, missed targets) that are not tied to a particular role.

\begin{figure}[t]
  \centering
  \includegraphics[width=0.95\linewidth]{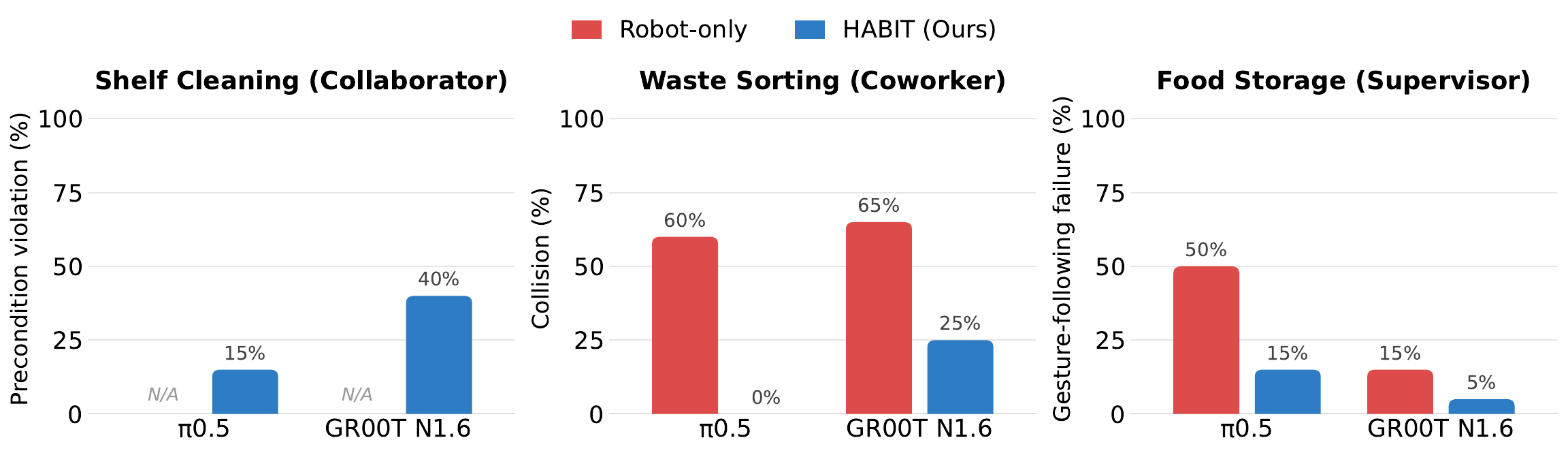}
  \caption{Role-specific failure analysis on one representative task per role. HABIT (Ours) substantially reduces precondition violations, collisions, and gesture-following failures compared to Robot-only baselines across both $\pi_{0.5}$ and GR00T N1.6. Robot-only is not applicable (N/A) for the Collaborator task.}
  \label{fig:failure-analysis}
  \vspace{-0.1in}
\end{figure}

Figure~\ref{fig:failure-analysis} reports role-specific failure rates for $\pi_{0.5}$ and GR00T N1.6 trained on Robot-only and \method~datasets. \method~substantially reduces role-specific failures across all three roles by instilling human-aware behaviors that conventional demonstration data cannot teach: synchronizing subtasks in Collaborator tasks, yielding to prevent collisions in Coworker tasks, and following human gestures in Supervisor tasks.

\subsection{Sample-Efficient Adaptation to New Tasks}
\label{sec:exp-scaling}

To investigate whether mid-training on \method~enables sample-efficient adaptation to downstream human-robot interaction tasks, we mid-train $\pi_{0.5}$ for 2 epochs on a subset of \method, sampling up to 100 demonstrations per task. 
The six evaluation tasks are excluded from this subset to prevent test-time leakage into the prior. 
We then compare $\pi_{0.5}$ with and without mid-training on Shelf Cleaning and Waste Sorting, fine-tuning each variant on 50, 100, and 200 demonstrations per task. 
Full implementation details are in Appendix~\ref{app:mid-training}.

\begin{figure}[t]
  \centering
  \includegraphics[width=0.9\linewidth]{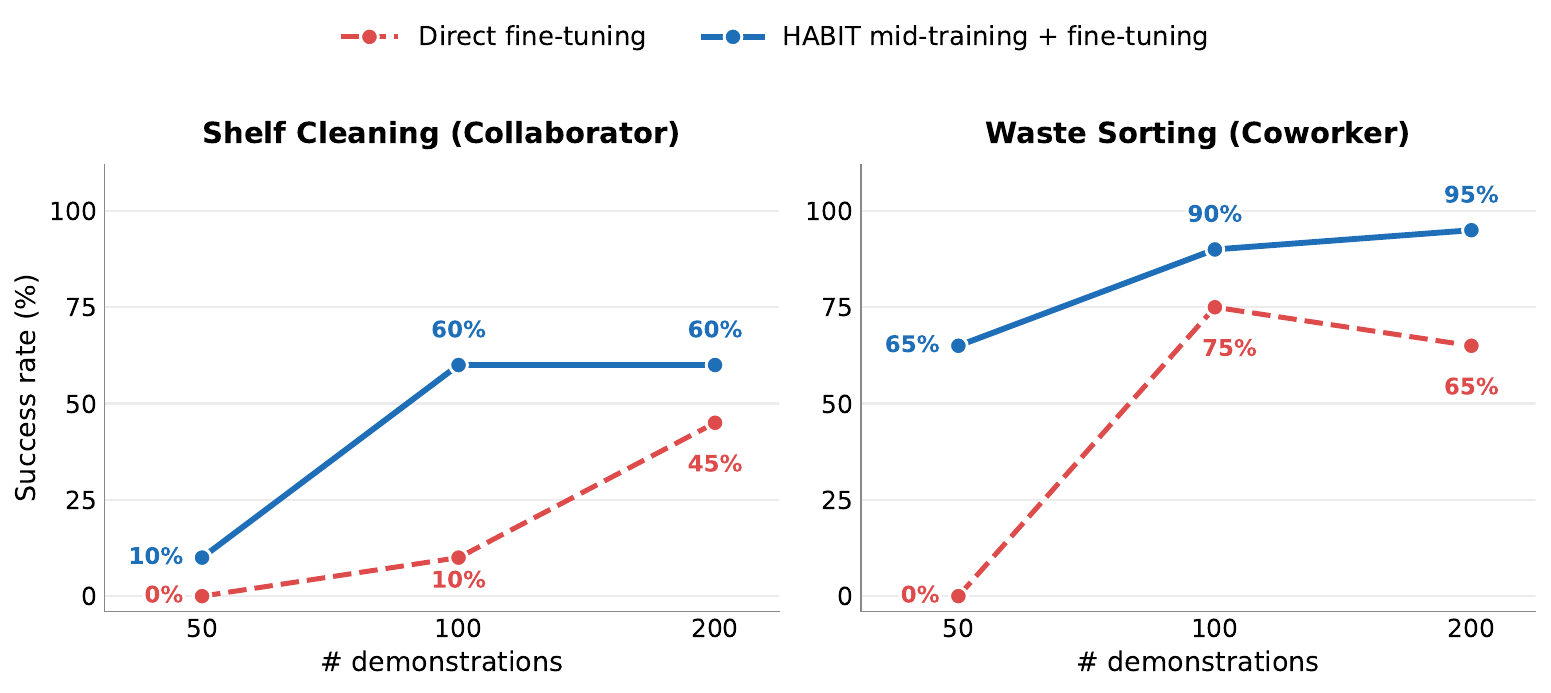}
  \caption{Sample efficiency of HABIT mid-training vs.\ direct fine-tuning. HABIT mid-training consistently achieves higher success rates with fewer demonstrations.}
  \label{fig:scaling}
  \vspace{-0.1in}
\end{figure}

As shown in Figure~\ref{fig:scaling}, mid-training on \method~improves both sample efficiency and final performance on downstream interaction tasks. On Waste Sorting, mid-training with 50 fine-tuning demonstrations matches direct fine-tuning at 200, and mid-training with 100 demonstrations surpasses direct fine-tuning at every tested budget. The effect is more pronounced on Shelf Cleaning, where direct fine-tuning fails to exceed a 45\% success rate even at 200 demonstrations, while mid-training with only 100 demonstrations reaches 60\%. 
These results indicate that \method~serves as a strong prior that transfers to new human-robot interaction tasks.
\section{Related Work}
\label{sec:related}

\paragraph{Large-scale robot manipulation datasets} 
Large-scale robot learning has evolved through a broad range of self-supervised, imitation-learning, multitask, and multi-robot datasets and systems~\citep{rt1,jang2022bc,robonet,pinto_gupta,mime,roboturk,qtopt,mtopt}. Recent work has further scaled robot manipulation datasets along several dimensions. DROID~\citep{droid}, BridgeData V2~\citep{walke2023bridgedata}, and RH20T~\citep{rh20t} emphasize diversity across scenes, tasks, and modalities to support broad generalization. Open X-Embodiment~\citep{open_x_embodiment} aggregates demonstrations across heterogeneous robot platforms to enable cross-embodiment transfer, while AgiBot World~\citep{agibot_world} and RoboCOIN~\citep{robocoin} scale bimanual manipulation on humanoid and multi-platform setups, respectively. Additional recent efforts further expand the scale and diversity of robot manipulation data~\citep{wu2024robomind,roboagent,open_galaxea}. Despite this breadth, these datasets share a common assumption: the robot is the only active agent in the workspace. As a result, they enable training manipulation policies for human-absent settings but provide no supervision for how a robot should behave when sharing its workspace with a co-present human. \method~addresses this gap by providing large-scale demonstrations in which the robot perceives and responds to an independent human partner.

\paragraph{Human-robot interaction/collaboration}
A complementary line of work in the Human-Robot Interaction (HRI) literature has long studied how robots and humans should share roles and tasks, with role taxonomies defining how a human and a robot relate to each other along axes such as goal sharing, task sharing, and team hierarchy \citep{scholtz2003, onnasch2021, pascher2023}. \method~builds on the role taxonomies from this prior HRI work, adopting the Collaborator, Coworker, and Supervisor distinction as the organizing structure of the dataset.
On the data side, prior HRI datasets that include paired robot actions fall short of true human-robot interaction.
In one line of work, the human directly teleoperates the robot as a tool to serve their own intent in a single assistive task \citep{harmonic}.
In another, the robot executes pre-scripted behaviors while the human reacts around it, with the data intended for safety risk monitoring rather than policy learning \citep{lihra}.
In neither case does the robot actively perceive and respond to an independent human. 
\method~fills this gap with large-scale, robot-action-paired manipulation data across diverse tasks, where the robot actively perceives and responds to an independent human under each of the three roles.
\section{Limitations}
\label{sec:limitations}
\method~has several limitations that bound the scope of its claims. The dataset is collected in a single environment with ten human operators under a one-to-one human-robot configuration, which limits the diversity of body silhouettes, motion styles, environmental conditions, and multi-agent dynamics a deployed robot would encounter. We partially counter this by systematically varying operator appearance and collection conditions across episodes, enabling the OOD analysis in Appendix~\ref{app:ood}. Beyond data, real-world evaluation is hard to reproduce. To support reproduction as closely as possible, we publish detailed per-task evaluation protocols (Appendix~\ref{app:task-details}).

\section{Conclusion and Future Directions}
\label{sec:conclusion}

We introduced \method, a large-scale robot demonstration dataset for human-present environments. Training two open-source VLAs on \method~consistently improves task success over robot-only baselines and elicits role-specific human-aware behaviors absent from robot-only data: reactive yielding, gesture grounding, and spatiotemporal synchronization. Furthermore, \method~enables sample-efficient adaptation to new human-robot interaction tasks. Co-training \method~with robot-only data shows a complementary effect, benefiting the policy both with and without a co-present human (Appendix~\ref{app:cotrain}). Together, these findings establish human presence as a learnable and impactful axis of data diversity for robot learning.

For future work, two directions naturally extend this work. First, although \method~releases five RGB streams per episode (three robot-side and two human-side), our experiments use only the three robot-side streams. Human egocentric view and the exocentric view plausibly carry signals that the robot-side cameras alone cannot recover, such as the precise referent of a pointing gesture~\citep{lee2026hierarchical} and the global spatial relation between the two agents. Incorporating these views into policy training is a promising direction for further improving human-aware behavior. Second, our experiments establish that mid-training on \method~transfers as a strong prior to downstream interaction tasks. Whether human-aware behavior emerges when this prior is fine-tuned on robot-only demonstrations of a new task is a promising direction for future work.

\begin{ack}
This work was supported by the Institute of Information \& Communications Technology Planning \& Evaluation (IITP) grant funded by the Korea government (MSIT) (RS-2019-II190075, Artificial Intelligence Graduate School Program (KAIST)); by the National Research Foundation of Korea (NRF) grant funded by the Korea government (MSIT) (RS-2026-25540546); and by the National Research Foundation of Korea (NRF) grant funded by the Korea government (MSIT) (No. RS-2026-25473123).
\end{ack}

\bibliographystyle{unsrtnat}
\bibliography{references}

@article{droid,
  author={Khazatsky, Alexander and Pertsch, Karl and Nair, Suraj and Balakrishna, Ashwin and Dasari, Sudeep and Karamcheti, Siddharth and Nasiriany, Soroush and Srirama, Mohan Kumar and Chen, Lawrence Yunliang and Ellis, Kirsty and others},
  title={{DROID}: A large-scale in-the-wild robot manipulation dataset},
  journal={arXiv preprint arXiv:2403.12945},
  year={2024}
}

@inproceedings{open_x_embodiment,
  author={O’Neill, Abby and Rehman, Abdul and Maddukuri, Abhiram and Gupta, Abhishek and Padalkar, Abhishek and Lee, Abraham and Pooley, Acorn and Gupta, Agrim and Mandlekar, Ajay and Jain, Ajinkya and others},
  title={{Open X-Embodiment}: Robotic learning datasets and rt-x models: Open x-embodiment collaboration 0},
  booktitle={International Conference on Robotics and Automation},
  year={2024}
}

@article{agibot_world,
  author={Bu, Qingwen and Cai, Jisong and Chen, Li and Cui, Xiuqi and Ding, Yan and Feng, Siyuan and Gao, Shenyuan and He, Xindong and Hu, Xuan and Huang, Xu and others},
  title={{AgiBot World Colosseo}: A large-scale manipulation platform for scalable and intelligent embodied systems},
  journal={arXiv preprint arXiv:2503.06669},
  year={2025}
}

@article{robocoin,
  author={Wu, Shihan and Liu, Xuecheng and Xie, Shaoxuan and Wang, Pengwei and Li, Xinghang and Yang, Bowen and Li, Zhe and Zhu, Kai and Wu, Hongyu and Liu, Yiheng and others},
  title={RoboCOIN: An Open-Sourced Bimanual Robotic Data COllection for INtegrated Manipulation},
  journal={arXiv preprint arXiv:2511.17441},
  year={2025}
}

@inproceedings{pinto_gupta,
  author={Pinto, Lerrel and Gupta, Abhinav},
  title={Supersizing self-supervision: Learning to grasp from 50k tries and 700 robot hours},
  booktitle={International conference on robotics and automation},
  year={2016}
}

@inproceedings{mime,
  author={Sharma, Pratyusha and Mohan, Lekha and Pinto, Lerrel and Gupta, Abhinav},
  title={Multiple interactions made easy (mime): Large scale demonstrations data for imitation},
  booktitle={Conference on robot learning},
  year={2018}
}

@inproceedings{roboturk,
  author={Mandlekar, Ajay and Zhu, Yuke and Garg, Animesh and Booher, Jonathan and Spero, Max and Tung, Albert and Gao, Julian and Emmons, John and Gupta, Anchit and Orbay, Emre and others},
  title={Roboturk: A crowdsourcing platform for robotic skill learning through imitation},
  booktitle={Conference on Robot Learning},
  year={2018}
}

@article{robonet,
  author={Dasari, Sudeep and Ebert, Frederik and Tian, Stephen and Nair, Suraj and Bucher, Bernadette and Schmeckpeper, Karl and Singh, Siddharth and Levine, Sergey and Finn, Chelsea},
  title={Robonet: Large-scale multi-robot learning},
  journal={arXiv preprint arXiv:1910.11215},
  year={2019}
}

@inproceedings{qtopt,
  author={Kalashnikov, Dmitry and Irpan, Alex and Pastor, Peter and Ibarz, Julian and Herzog, Alexander and Jang, Eric and Quillen, Deirdre and Holly, Ethan and Kalakrishnan, Mrinal and Vanhoucke, Vincent and others},
  title={Scalable deep reinforcement learning for vision-based robotic manipulation},
  booktitle={Conference on robot learning},
  year={2018}
}

@article{mtopt,
  author={Kalashnikov, Dmitry and Varley, Jacob and Chebotar, Yevgen and Swanson, Benjamin and Jonschkowski, Rico and Finn, Chelsea and Levine, Sergey and Hausman, Karol},
  title={Mt-opt: Continuous multi-task robotic reinforcement learning at scale},
  journal={arXiv preprint arXiv:2104.08212},
  year={2021}
}

@article{rt1,
  author={Brohan, Anthony and Brown, Noah and Carbajal, Justice and Chebotar, Yevgen and Dabis, Joseph and Finn, Chelsea and Gopalakrishnan, Keerthana and Hausman, Karol and Herzog, Alex and Hsu, Jasmine and others},
  title={Rt-1: Robotics transformer for real-world control at scale},
  journal={arXiv preprint arXiv:2212.06817},
  year={2022}
}

@inproceedings{rh20t,
  author={Fang, Hao-Shu and Fang, Hongjie and Tang, Zhenyu and Liu, Jirong and Wang, Chenxi and Wang, Junbo and Zhu, Haoyi and Lu, Cewu},
  title={Rh20t: A comprehensive robotic dataset for learning diverse skills in one-shot},
  booktitle={International Conference on Robotics and Automation},
  year={2024}
}

@inproceedings{roboagent,
  author={Bharadhwaj, Homanga and Vakil, Jay and Sharma, Mohit and Gupta, Abhinav and Tulsiani, Shubham and Kumar, Vikash},
  title={Roboagent: Generalization and efficiency in robot manipulation via semantic augmentations and action chunking},
  booktitle={International Conference on Robotics and Automation},
  year={2024}
}

@article{open_galaxea,
  author={Jiang, Tao and Yuan, Tianyuan and Liu, Yicheng and Lu, Chenhao and Cui, Jianning and Liu, Xiao and Cheng, Shuiqi and Gao, Jiyang and Xu, Huazhe and Zhao, Hang},
  title={Galaxea open-world dataset and g0 dual-system vla model},
  journal={arXiv preprint arXiv:2509.00576},
  year={2025}
}

@inproceedings{scholtz2003,
  author={Scholtz, Jean},
  title={Theory and evaluation of human robot interactions},
  booktitle={Hawaii International Conference on System Sciences},
  year={2003}
}

@article{onnasch2021,
  author={Onnasch, Linda and Roesler, Eileen},
  journal={International Journal of Social Robotics},
  number={4},
  pages={833--849},
  title={A taxonomy to structure and analyze human--robot interaction},
  volume={13},
  year={2021}
}

@inproceedings{pascher2023,
  author={Pascher, Max and Gruenefeld, Uwe and Schneegass, Stefan and Gerken, Jens},
  title={How to communicate robot motion intent: A scoping review},
  booktitle={Conference on Human Factors in Computing Systems},
  year={2023}
}

@article{harmonic,
  author={Newman, Benjamin A and Aronson, Reuben M and Srinivasa, Siddhartha S and Kitani, Kris and Admoni, Henny},
  journal={The International Journal of Robotics Research},
  number={1},
  pages={3--11},
  title={HARMONIC: A multimodal dataset of assistive human--robot collaboration},
  volume={41},
  year={2022}
}

@inproceedings{lihra,
  author={Plahl, Frederik and Katranis, Georgios and Mamaev, Ilshat and Morozov, Andrey},
  title={LiHRA: A LiDAR-Based HRI Dataset for Automated Risk Monitoring Methods},
  booktitle={IEEE/RSJ International Conference on Intelligent Robots and Systems},
  year={2025}
}

@article{punamiya2026egoverse,
  author={Punamiya, Ryan and Kareer, Simar and Liu, Zeyi and Citron, Josh and Qiu, Ri-Zhao and Cai, Xiongyi and Gavryushin, Alexey and Chen, Jiaqi and Liconti, Davide and Zhu, Lawrence Y and others},
  title={EgoVerse: An Egocentric Human Dataset for Robot Learning from Around the World},
  journal={arXiv preprint arXiv:2604.07607},
  year={2026}
}

@inproceedings{ego4d,
  author={Grauman, Kristen and Westbury, Andrew and Byrne, Eugene and Chavis, Zachary and Furnari, Antonino and Girdhar, Rohit and Hamburger, Jackson and Jiang, Hao and Liu, Miao and Liu, Xingyu and others},
  title={{Ego4D}: Around the world in 3,000 hours of egocentric video},
  booktitle={IEEE/CVF conference on computer vision and pattern recognition},
  year={2022}
}

@inproceedings{ego_exo4d,
  author={Grauman, Kristen and Westbury, Andrew and Torresani, Lorenzo and Kitani, Kris and Malik, Jitendra and Afouras, Triantafyllos and Ashutosh, Kumar and Baiyya, Vijay and Bansal, Siddhant and Boote, Bikram and others},
  title={{Ego-Exo4D}: Understanding skilled human activity from first-and third-person perspectives},
  booktitle={IEEE/CVF Conference on Computer Vision and Pattern Recognition},
  year={2024}
}

@article{pi05,
  author={{Physical Intelligence} and Black, Kevin and Brown, Noah and Darpinian, James and Dhabalia, Karan and Driess, Danny and Esmail, Adnan and Equi, Michael and Finn, Chelsea and Fusai, Niccolo and others},
  title={{$\pi_{0.5}$}: A Vision-Language-Action Model with Open-World Generalization},
  journal={arXiv preprint arXiv:2504.16054},
  year={2025}
}

@article{pi0,
  author={Black, Kevin and Brown, Noah and Driess, Danny and Esmail, Adnan and Equi, Michael and Finn, Chelsea and Fusai, Niccolo and Groom, Lachy and Hausman, Karol and Ichter, Brian and others},
  title={{$\pi_0$}: A Vision-Language-Action Flow Model for General Robot Control},
  journal={arXiv preprint arXiv:2410.24164},
  year={2024}
}

@article{groot,
  author={Bjorck, Johan and Casta{\~n}eda, Fernando and Cherniadev, Nikita and Da, Xingye and Ding, Runyu and Fan, Linxi and Fang, Yu and Fox, Dieter and Hu, Fengyuan and Huang, Spencer and others},
  title={{GR00T N1}: An open foundation model for generalist humanoid robots},
  journal={arXiv preprint arXiv:2503.14734},
  year={2025}
}

@inproceedings{rt2,
  author={Zitkovich, Brianna and Yu, Tianhe and Xu, Sichun and Xu, Peng and Xiao, Ted and Xia, Fei and Wu, Jialin and Wohlhart, Paul and Welker, Stefan and Wahid, Ayzaan and others},
  title={{RT-2}: Vision-language-action models transfer web knowledge to robotic control},
  booktitle={Conference on Robot Learning},
  year={2023}
}

@article{openvla,
  author={Kim, Moo Jin and Pertsch, Karl and Karamcheti, Siddharth and Xiao, Ted and Balakrishna, Ashwin and Nair, Suraj and Rafailov, Rafael and Foster, Ethan and Lam, Grace and Sanketi, Pannag and others},
  title={{OpenVLA}: An open-source vision-language-action model},
  journal={arXiv preprint arXiv:2406.09246},
  year={2024}
}

@article{octo,
  author={{Octo Model Team} and Ghosh, Dibya and Walke, Homer and Pertsch, Karl and Black, Kevin and Mees, Oier and Dasari, Sudeep and Hejna, Joey and Kreiman, Tobias and Xu, Charles and others},
  title={{Octo}: An open-source generalist robot policy},
  journal={arXiv preprint arXiv:2405.12213},
  year={2024}
}

@article{dreamzero,
  author={Ye, Seonghyeon and Ge, Yunhao and Zheng, Kaiyuan and Gao, Shenyuan and Yu, Sihyun and Kurian, George and Indupuru, Suneel and Tan, You Liang and Zhu, Chuning and Xiang, Jiannan and others},
  title={World action models are zero-shot policies},
  journal={arXiv preprint arXiv:2602.15922},
  year={2026}
}

@article{cosmospolicy,
  author={Kim, Moo Jin and Gao, Yihuai and Lin, Tsung-Yi and Lin, Yen-Chen and Ge, Yunhao and Lam, Grace and Liang, Percy and Song, Shuran and Liu, Ming-Yu and Finn, Chelsea and others},
  title={{Cosmos Policy}: Fine-tuning video models for visuomotor control and planning},
  journal={arXiv preprint arXiv:2601.16163},
  year={2026}
}

@article{mimicvideo,
  author={Pai, Jonas and Achenbach, Liam and Montesinos, Victoriano and Forrai, Benedek and Mees, Oier and Nava, Elvis},
  title={{mimic-video}: Video-action models for generalizable robot control beyond vlas},
  journal={arXiv preprint arXiv:2512.15692},
  year={2025}
}

@article{liang2025video,
  author={Liang, Junbang and Tokmakov, Pavel and Liu, Ruoshi and Sudhakar, Sruthi and Shah, Paarth and Ambrus, Rares and Vondrick, Carl},
  title={Video generators are robot policies},
  journal={arXiv preprint arXiv:2508.00795},
  year={2025}
}

@inproceedings{jang2022bc,
  author={Jang, Eric and Irpan, Alex and Khansari, Mohi and Kappler, Daniel and Ebert, Frederik and Lynch, Corey and Levine, Sergey and Finn, Chelsea},
  title={Bc-z: Zero-shot task generalization with robotic imitation learning},
  booktitle={conference on Robot Learning},
  year={2022},
}

@inproceedings{walke2023bridgedata,
  author={Walke, Homer Rich and Black, Kevin and Zhao, Tony Z and Vuong, Quan and Zheng, Chongyi and Hansen-Estruch, Philippe and He, Andre Wang and Myers, Vivek and Kim, Moo Jin and Du, Max and others},
  title={Bridgedata v2: A dataset for robot learning at scale},
  booktitle={Conference on Robot Learning},
  year={2023},
}

@article{wu2024robomind,
  author={Wu, Kun and Hou, Chengkai and Liu, Jiaming and Che, Zhengping and Ju, Xiaozhu and Yang, Zhuqin and Li, Meng and Zhao, Yinuo and Xu, Zhiyuan and Yang, Guang and others},
  title={Robomind: Benchmark on multi-embodiment intelligence normative data for robot manipulation},
  journal={arXiv preprint arXiv:2412.13877},
  year={2024}
}

@article{bauer2008human,
  author={Bauer, Andrea and Wollherr, Dirk and Buss, Martin},
  journal={International Journal of Humanoid Robotics},
  number={01},
  pages={47--66},
  title={Human--robot collaboration: a survey},
  volume={5},
  year={2008},
}

@article{zheng2026egoscale,
  author={Zheng, Ruijie and Niu, Dantong and Xie, Yuqi and Wang, Jing and Xu, Mengda and Jiang, Yunfan and Casta{\~n}eda, Fernando and Hu, Fengyuan and Tan, You Liang and Fu, Letian and others},
  title={Egoscale: Scaling dexterous manipulation with diverse egocentric human data},
  journal={arXiv preprint arXiv:2602.16710},
  year={2026}
}

@inproceedings{de2019causal,
 author={De Haan, Pim and Jayaraman, Dinesh and Levine, Sergey},
 title={Causal confusion in imitation learning},
 booktitle={Advances in Neural Information Processing Systems},
 year={2019}
}

@article{xu2026learning,
  author={Xu, Leo and Li, Letian and Cuellar, Alex and Hagenow, Michael},
  title={Learning to Assist: Collaborative VLAs for Implicit Human-Robot Collaboration},
  journal={arXiv preprint arXiv:2606.12475},
  year={2026}
}

@article{lee2026hierarchical,
  author  = {Dongjun Lee and Juheon Choi and Dong Kyu Shin and Sinjae Kang and Kimin Lee},
  title   = {EDITH: Teaching Robot to Follow Nonverbal Human Intent from Egocentric Vision and Gaze},
  journal = {arXiv preprint arXiv:2606.10276},
  year    = {2026}
}

\newpage
\appendix
\begin{center}{\bf {\LARGE Appendix:}}
\end{center}
\begin{center}{\bf {\Large HABIT: Human-Aware Behavior and Interaction Training Dataset for Robot Manipulation}}
\end{center}

\begin{figure}[h]
\centering
\includegraphics[width=\linewidth]{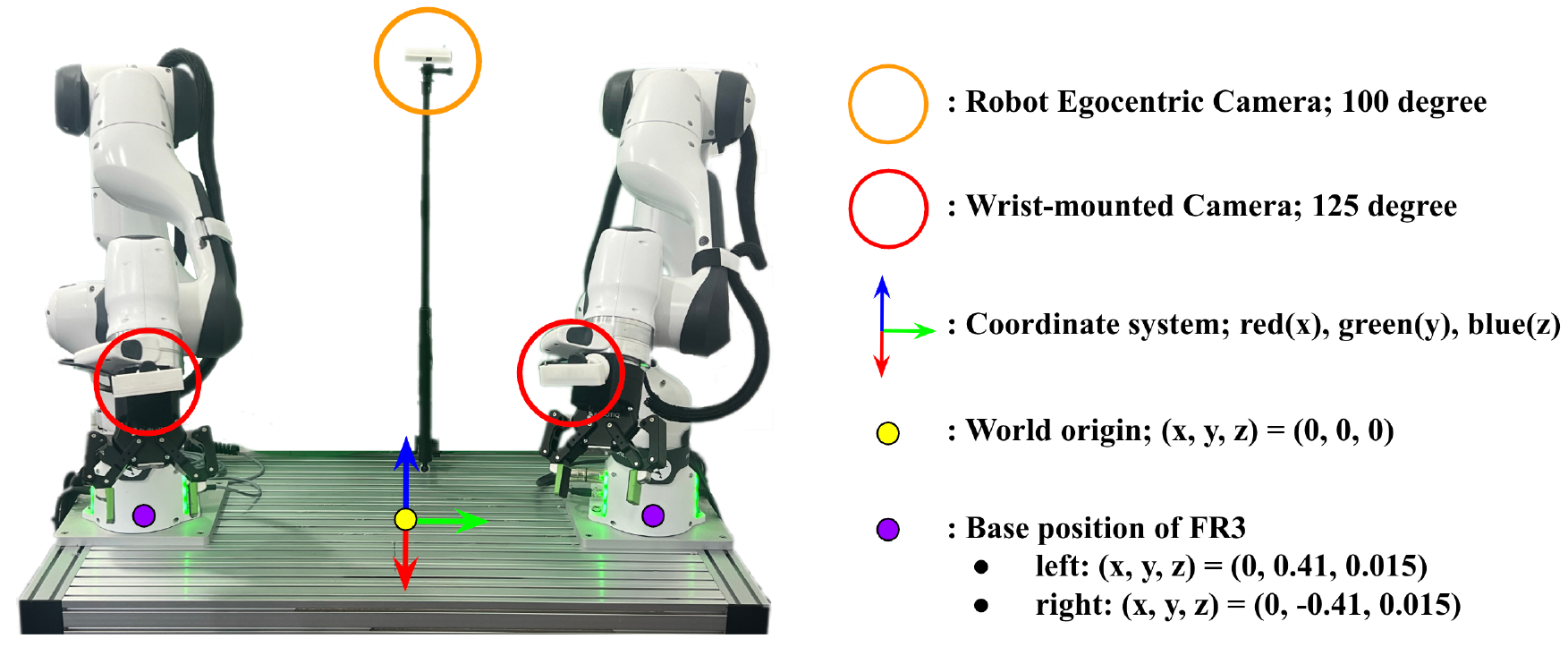}
\caption{
Robot-side workspace detail.
The world origin (yellow) lies at the midpoint between the two Franka Research 3 (FR3) base centers (purple) on the frame surface, with axes $+x$ forward, $+y$ left, and $+z$ up.
The orange circle marks the center camera ($100^\circ$ FoV), while the red circles mark the two wrist-mounted cameras ($125^\circ$ FoV).
}
\label{fig:robot_setup}
\end{figure}

\section{Hardware and Collection Details}
\label{app:hardware}

\subsection{Workspace and Coordinate System}
\label{app:workspace}

Figure~\ref{fig:robot_setup} shows the robot-side workspace in detail.
We define a shared world frame whose origin lies at the midpoint between the two arm bases on the surface of the aluminum frame.
The axes follow the convention $+x$ forward (away from the robots and toward the human), $+y$ to the left, and $+z$ upward.
All translations are reported in meters and all rotations in radians using Euler angles.
Cartesian end-effector poses stored in the released parquet files are expressed in each arm's own base frame, and joint-space actions are likewise defined per arm.

The two Franka Research~3 (FR3) arms are mounted symmetrically about the world origin.
Their base positions are $(0,\,+0.41,\,0.015)\,\text{m}$ for the left arm and $(0,\,-0.41,\,0.015)\,\text{m}$ for the right arm, giving a base-to-base separation of $0.82$~m.
The base mounting plate sits $0.765$~m above the floor ($0.75$~m table height plus a $0.015$~m mounting plate).
Each arm is equipped with a Robotiq parallel-jaw gripper.
Both arms use a fixed initial joint configuration across all episodes:
$\mathbf{q}_{\text{left}} =
[-0.3507,\,-0.2842,\,-0.0117,\,-2.7405,\,0.0340,\,3.0094,\,0.2150]$
rad and
$\mathbf{q}_{\text{right}} =
[0.3435,\,-0.2832,\,0.0739,\,-2.7468,\,0.0666,\,3.0379,\,-0.3361]$
rad.

\subsection{Teleoperation}
\label{app:teleop}
The teleoperation pipeline is based on the DROID codebase~\citep{droid}, with commands specifying desired joint positions, velocities, and accelerations.
Operators control the two arms via the hand controllers of a Meta Quest~3 headset.
Action representations are recorded in joint-space, Cartesian-space, and gripper-state form so that downstream users may train policies in whichever action space their model expects.

\subsection{Camera Specifications}
\label{app:cameras}
We record synchronized RGB streams from five cameras.
On the robot side, two wrist-mounted cameras (one per arm) provide close-up views of each gripper with a $125^\circ$ field of view, and a center camera mounted on a pole between the arms provides a forward-facing egocentric view of both the human and the shared workspace at $100^\circ$ field of view (Figure~\ref{fig:robot_setup}).
On the human side, a head-mounted camera captures the operator's egocentric perspective, and a fifth camera positioned off to the side captures a holistic third-person view of the full human-robot interaction.
All cameras are software-synchronized and recorded at 640×480 resolution and 10 Hz.

\subsection{Analysis of Data Collection Protocol}
\label{app:protocol}
Each component of the collection protocol in Section~\ref{sec:data-collection} was introduced in response to observations made during development. In an early version of Waste Sorting, robot operators paused in place rather than retreating when the human approached, and the trained policy frequently collided with the human. Once operators were instructed to retreat, the policy learned to reactively avoid the approaching human, which led us to adopt safety-prioritized yielding. Similarly, when the human pointed immediately after the instruction was given, the trained policy began acting before the gesture appeared. A related form of premature assistance has also been reported in collaborative VLA policies~\citep{xu2026learning}. Randomizing the wait time before pointing removed this premature behavior, and the resulting policy waits stably for gestures delayed by up to four minutes, far beyond the 30-second maximum in the dataset. Finally, if operators acted on their prior knowledge of the workflow rather than on what they observed, the cue that triggers each action would not appear in the recorded observations, leaving the policy with no learnable signal for when to act and potentially inducing causal confusion~\citep{de2019causal}. We adopt reactive interaction to prevent this.

These observations indicate that human presence alone is not sufficient. Human-aware behaviors emerge when they are consistently demonstrated in the data, and each protocol component was introduced to ensure such demonstrations. We therefore attribute the gains of \method~over Robot-only data to the human-present, behavior-eliciting collection protocol as a whole rather than to human presence alone.


\section{Object Diversity}
\label{app:objects}
We summarize the statistics of the objects used across the tasks in \method. Objects are counted at the type level, so instances of the same type that differ in color or shape are counted once, while each type itself appears in a variety of colors and shapes. As summarized in Table~\ref{tab:objects}, \method~involves 101 distinct object types across 12 categories, ranging from kitchenware and containers to electronics and labware. Each task uses on average 4.2 object types, with a median of 4 and a range of 2 to 8.

Objects are shared across multiple scenarios and roles rather than tied to a single task: 59 of the 101 types appear in two or more tasks, and 54 appear under two or more interaction roles. This reflects our design goal of capturing the same objects across different modes of human-robot interaction. The full per-task object list is available on the dataset page.

\begin{table}[h]
\centering
\small
\caption{Object categories in \method. Counts denote the number of distinct object types per category.}
\label{tab:objects}
\begin{tabular}{l c l}
\toprule
Category & \# Types & Example objects \\
\midrule
Kitchenware & 17 & picnic bowl, reusable cup, tongs, baking sheet \\
Containers & 13 & organizing basket, airtight container, laundry basket \\
Stationery & 12 & stapler, pencil case, highlighter, box cutter \\
Serving equipment & 9 & handle tray, can dispenser, bakery case \\
Food \& beverages & 9 & donut, butter roll, beverage cans, bottled water \\
Packaging & 9 & mailer box, bubble wrap, paper tape, to-go box \\
Electronics & 8 & PCB, PCB rack, power cable, digital clock \\
Labware & 7 & flask, measuring cup, funnel, lab tube \\
Textiles \& clothing & 6 & towel, napkin, shoes, clothing \\
Cleaning & 5 & duster, lint roller, shoe brush \\
Furniture & 4 & shelf, serving table, shoe stand \\
Accessories \& toys & 2 & glasses, plush toy \\
\midrule
Total & 101 & \\
\bottomrule
\end{tabular}
\end{table}

\section{Evaluation Task Details}
\label{app:task-details}

This section provides full setup, evaluation criteria, evaluation protocols, and task workflow for the six evaluation tasks introduced in Section~\ref{sec:evaluation}. The two tasks per role vary along an axis that stresses each role's characteristic requirement, summarized below and detailed in the subsections.

\begin{itemize} [leftmargin=1.5em, itemsep=1mm, topsep=1pt]
    \item \textbf{Collaborator (spatiotemporal synchronization)}: \emph{Table Serving} and \emph{Shelf Cleaning} both require the robot's action to align with the human's ongoing action both spatially and temporally.
    \item \textbf{Coworker (reactive collision avoidance)}: \emph{Waste Sorting} (moderate workspace overlap) and \emph{Box Packaging} (high overlap) differ in how often the robot's manipulation path conflicts with the human's.
    \item \textbf{Supervisor (gesture following)}: \emph{Donut Serving} (single-stage trajectory) and \emph{Food Storage} (two-stage trajectory) differ in trajectory structure while sharing the same gesture-grounding requirement.
\end{itemize}

\subsection{Task Setup and Randomization}
\label{app:task-setup}

\paragraph{Table Serving (Collaborator).}
Two trays are pre-set with a cup and bowl on the robot's table. The human approaches one tray with a folded napkin, and the robot must lift the dishware (cup and bowl) from that specific tray and hold it in the air while the human lays the napkin underneath, then place the dishware back on top of the napkin. The process repeats for the second tray. We randomize which tray the human approaches first, the left/right arrangement of cup and bowl on each tray, and the cup/bowl colors.

\paragraph{Shelf Cleaning (Collaborator).}
The task is bracketed by a duster handover at both ends. The human first hands the robot a duster, then lifts the objects off one tier of a 2-tier shelf (chosen randomly) so the robot can dust that tier, then lifts objects off the next tier, and finally receives the duster back after the robot has cleaned the last tier. Paper confetti is sprinkled on each tier as the visible target of cleaning. We randomize the hand (left/right) used for the duster handover, the order in which the human clears tiers, and the placement and type of objects on each tier.

\paragraph{Waste Sorting (Coworker).}
Cans, glass bottles, and plastic bottles are scattered on the table, with three labeled bins on the robot's side. The robot sorts cans only, while the human sorts glass and plastic bottles in parallel into the same set of bins. Both agents reach into the central area to pick up items, but the robot retreats to its own side to place them, so human-robot workspace overlap is \emph{moderate} (averaging 2 yielding events per trial). We randomize the positions of all items on the table, and pre-specify the order in which the human picks up items so that evaluation is reproducible across trials.

\paragraph{Box Packaging (Coworker).}
Two mailer boxes are placed at the table's mid-left, and stationery items are scattered in the middle and right of the table. The robot and human each pack their own box (one pencil case and one stapler each, for the robot) and close the lid. Because the boxes are at the table's center-left and items must traverse this region from both sides, human-robot workspace overlap is \emph{high} (averaging 3 yielding events per trial). We randomize the positions of stationery items, and pre-specify the order in which the human picks up items so that evaluation is reproducible across trials.

\paragraph{Donut Serving (Supervisor).}
Two roll-top bakery cases each contain two donuts in to-go boxes (four donuts total). A tray sits at the table's mid-bottom. The human points to one donut, and the robot must lift the corresponding to-go box (with donut inside) onto the tray. The high-level instruction takes the form ``place the $k$-th donut from the left on the tray'', with $k$ sampled uniformly across the four positions in both training and evaluation. The trajectory is single-stage (direct from the bakery case to the tray). We randomize $k$ and apply small perturbations to the donut arrangement within the bakery cases.

\paragraph{Food Storage (Supervisor).}
Four open airtight containers are aligned at the top of the table, and one bread roll sits on a plate at the bottom. The human points to one of the four containers, and the robot must pick up the bread first and place it into the indicated container. The instruction takes the form ``place the bread in the $k$-th container from the left'', with $k$ sampled uniformly across the four positions. The trajectory is two-stage (pick up the bread, then place into the indicated container). We randomize $k$ and the bread's position and orientation on the plate (e.g., centered or partially overhanging the edge).

\paragraph{Wait time variation for Supervisor tasks.}
To prevent policies from short-cutting to action immediately after the language instruction is issued, the human's wait time before pointing is deliberately varied during data collection. Wait times are drawn from bins of $\{0, 1\text{--}5, 5\text{--}10, 10\text{--}20, 20\text{--}30\}$ seconds for Food Storage and $\{0, 1\text{--}5, 5\text{--}10, 10\text{--}15, 15\text{--}20\}$ seconds for Donut Serving, with equal numbers of episodes per bin. This forces policies to ground the gesture before acting rather than relying on the language instruction alone.

\subsection{Per-Task Evaluation Criteria}
\label{app:eval-criteria}

For all six tasks, manipulation success requires completing the manipulation specified in the low-level instruction, and safety registers a violation on any human-robot collision during the trial. Workflow compliance is task-specific and follows the definitions of Section~\ref{sec:evaluation}.

\paragraph{Table Serving and Shelf Cleaning (Collaborator).}
The robot must act \emph{spatially} on the tray or tier the human is currently attending to, and \emph{temporally} only within the appropriate window of the human's action. Acting on the wrong tray/tier or acting before the human is ready (or after the human has moved on) registers a precondition violation, even when manipulation itself succeeds.

\paragraph{Waste Sorting and Box Packaging (Coworker).}
The robot's manipulation targets are independent of the human, so workflow compliance is trivially satisfied. The role-specific failure mode is captured through safety, which registers a violation whenever the robot collides with the human while reaching for or placing items in the shared workspace.

\paragraph{Food Storage and Donut Serving (Supervisor).}
The robot must place the bread in the container, or pick up the donut, indicated by the human's pointing gesture. Acting on the wrong target registers a gesture-following failure, even when the manipulation itself succeeds.

\subsection{Per-Task Evaluation Protocol}
\label{app:eval-protocol}

This section describes the detailed evaluation protocol used for each task, including object placement procedures, human action sequences, and success criteria. The protocol is designed to be reproducible across trials within an evaluation cell.

\begin{figure}[ht]
    \centering
    \includegraphics[width=\linewidth]{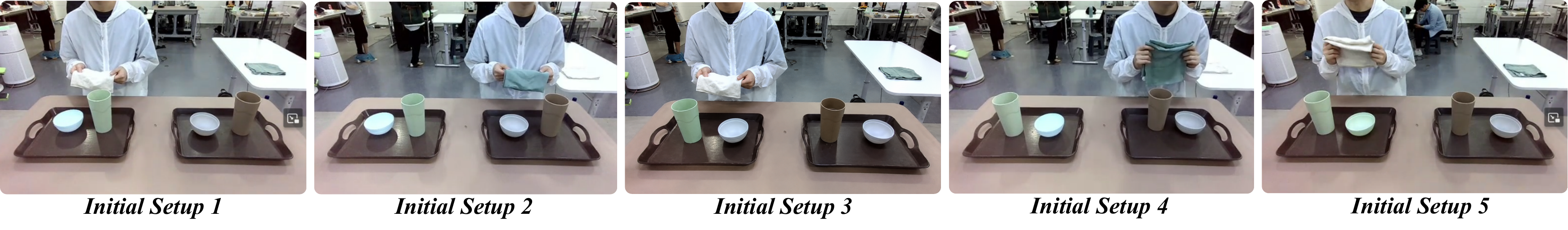}
    \caption{The five initial setups for the Table Serving task. The setups vary the cup-bowl arrangement on the trays, the color of the right-side bowl (changed only in the fifth setup), and the tray that the human operator approaches first with the napkin.}
    \label{fig:eval-table-serving}
\end{figure}

\paragraph{Table Serving.}
This task is evaluated over the 5 initial setups shown in Figure~\ref{fig:eval-table-serving}, each repeated 4 times for 20 trials total. The setups differ in three aspects: (i) the left-to-right arrangement of the cup and bowl on each tray, (ii) the color of the right-side bowl, which is changed only in the fifth setup, and (iii) the tray that the human operator approaches first with the napkin. For the in-distribution and OOD-silhouette evaluations, the human operator wears white. For the OOD-clothes evaluation, two clothing colors absent from the training data are used: pink and purple, with 10 trials each.

\begin{figure}[ht]
    \centering
    \includegraphics[width=\linewidth]{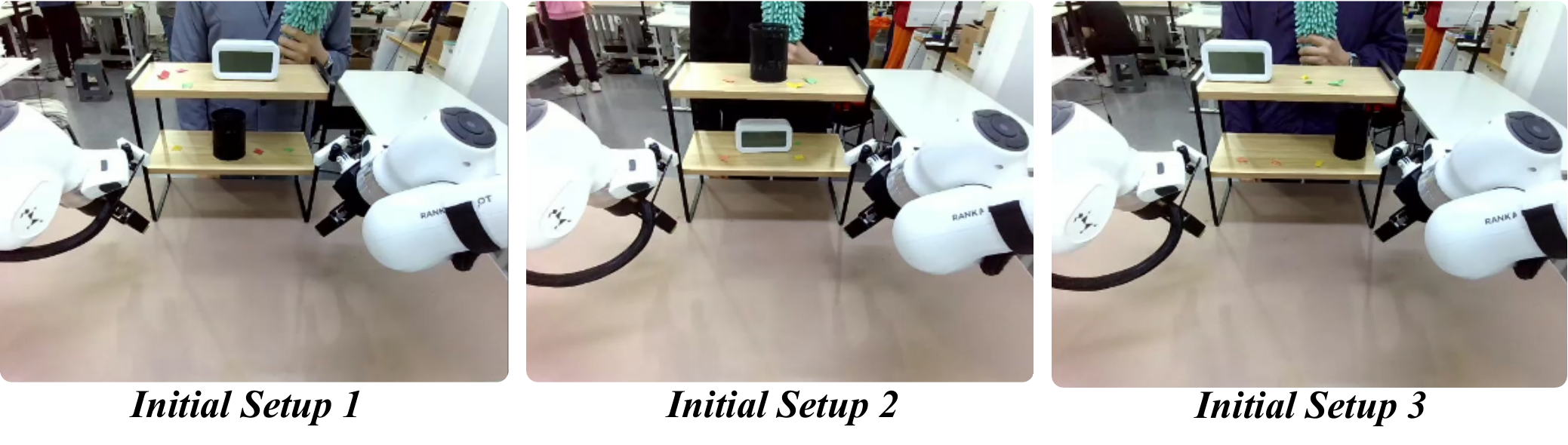}
    \caption{The three initial setups for the Shelf Cleaning task. The setups vary which object (clock or pencil case) is placed on each tier of the two-tier shelf and the position of each object on its tier.}
    \label{fig:eval-shelf-cleaning}
\end{figure}

\paragraph{Shelf Cleaning.}
This task is evaluated over the 3 initial setups shown in Figure~\ref{fig:eval-shelf-cleaning}, with 20 trials in total. In each trial, the human operator lifts an object from one tier of the shelf, and the robot cleans that tier with a duster received from the human. The setups place a clock on one tier and a pencil case on the other, varying the assignment and the position of each object on its tier. Within each setup, trials vary along two axes: which arm of the robot receives the duster (left or right), and which tier the human lifts an object from first (the upper or lower shelf). The first two setups cover all four combinations with 2 trials each, for 8 trials per setup, while the third setup covers only the upper-shelf-first case with 2 trials per arm, for 4 trials. For the in-distribution and OOD-silhouette evaluations, the human operator wears gray. For the OOD-clothes evaluation, two clothing colors absent from the training data are used: purple and black, with 10 trials each.

\begin{figure}[ht]
    \centering
    \includegraphics[width=\linewidth]{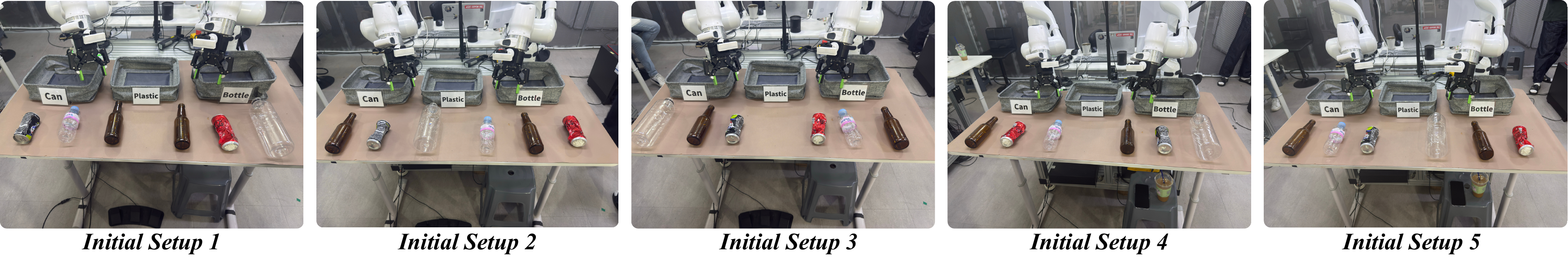}
    \caption{The five initial setups for the Waste Sorting task. The setups vary the arrangement of the four non-can objects on the table, with each setup using four distinct pickup orders (one per trial).}
    \label{fig:eval-waste-sorting}
\end{figure}

\paragraph{Waste Sorting.}
This task is evaluated over the 5 initial setups shown in Figure~\ref{fig:eval-waste-sorting}, each repeated 4 times for 20 trials total. Each setup specifies the arrangement of the four non-can objects on the table that the human operator picks up. Within each setup, the 4 trials use 4 distinct pickup orders, each a permutation of the four object indices (numbered 1 through 4 from the human operator's left):
\begin{itemize}[leftmargin=1.5em, itemsep=1mm, topsep=1pt]
    \item Initial setup 1: 2143, 1234, 3241, 1342
    \item Initial setup 2: 1324, 4321, 4123, 2314
    \item Initial setup 3: 1324, 3142, 3214, 3412
    \item Initial setup 4: 1234, 1243, 4231, 4321
    \item Initial setup 5: 3142, 1423, 4231, 2134
\end{itemize}
For the in-distribution and OOD-silhouette evaluations, the human operator wears black. For the OOD-clothes evaluation, two clothing colors absent from the training data are used: sky blue and orange, with 10 trials each.

\begin{figure}[ht]
    \centering
    \includegraphics[width=\linewidth]{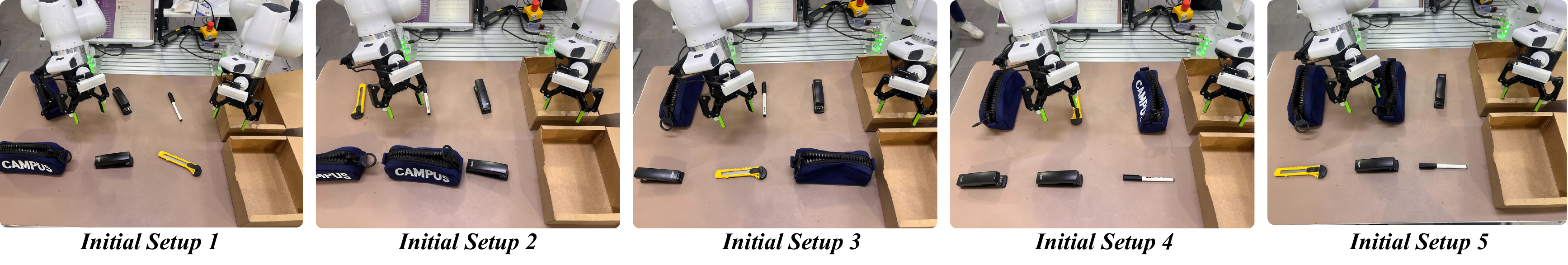}
    \caption{The five initial setups for the Box Packaging task. The setups vary the arrangement of the six stationery items (one knife, two pencil cases, one name pen, and two staplers) across the upper and lower rows, with each setup using two distinct pickup orders (each repeated twice).}
    \label{fig:eval-box-packaging}
\end{figure}

\paragraph{Box Packaging.}
This task is evaluated over the 5 initial setups shown in Figure~\ref{fig:eval-box-packaging}, with 2 distinct pickup orders per setup, each repeated twice for 20 trials total. The task involves packing six stationery items (one knife, two pencil cases, one name pen, and two staplers) into two boxes, one packed by the robot and one by the human. The robot packs only a pencil case and a stapler into its box, while the human packs the remaining four items, namely a pencil case, a stapler, the knife, and the name pen, into the other box. Each setup specifies the arrangement of the items across the upper and lower rows from the human operator's perspective. The 2 pickup orders per setup specify the row from which the human picks up each item, with U denoting the upper row and L denoting the lower row:
\begin{itemize}[leftmargin=1.5em, itemsep=1mm, topsep=1pt]
    \item Initial setup 1, order A: L-knife, L-pencil case, U-name pen, L-stapler
    \item Initial setup 1, order B: U-stapler, L-knife, U-name pen, L-pencil case
    \item Initial setup 2, order A: L-pencil case, L-stapler, U-name pen, U-knife
    \item Initial setup 2, order B: U-name pen, L-stapler, L-pencil case, U-knife
    \item Initial setup 3, order A: L-pencil case, L-stapler, L-knife, U-name pen
    \item Initial setup 3, order B: U-name pen, L-knife, L-pencil case, L-stapler
    \item Initial setup 4, order A: L-stapler, U-pencil case, L-name pen, U-knife
    \item Initial setup 4, order B: L-name pen, L-stapler, U-knife, U-pencil case
    \item Initial setup 5, order A: U-stapler, L-knife, U-name pen, U-pencil case
    \item Initial setup 5, order B: L-knife, L-stapler, L-name pen, U-pencil case
\end{itemize}
For the in-distribution and OOD-silhouette evaluations, the human operator wears blue. For the OOD-clothes evaluation, two clothing colors absent from the training data are used: purple and orange, with 10 trials each.

\begin{figure}[ht]
    \centering
    \begin{subfigure}{\linewidth}
        \centering
        \includegraphics[width=\linewidth]{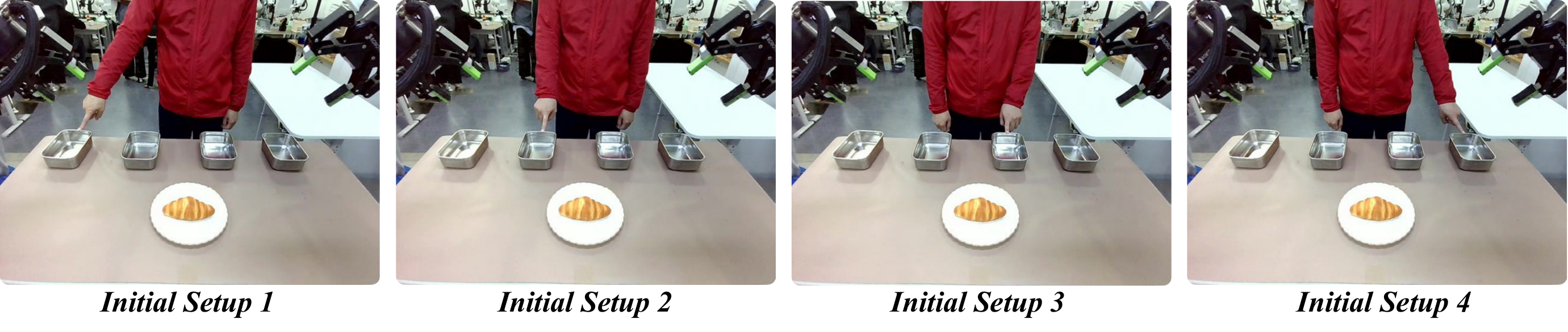}
        \caption{Pointing positions: the four containers indexed 1 through 4 from the human operator's right.}
        \label{fig:eval-food-storage-pointing}
    \end{subfigure}
    
    \vspace{0.5em}
    
    \begin{subfigure}{\linewidth}
        \centering
        \includegraphics[width=\linewidth]{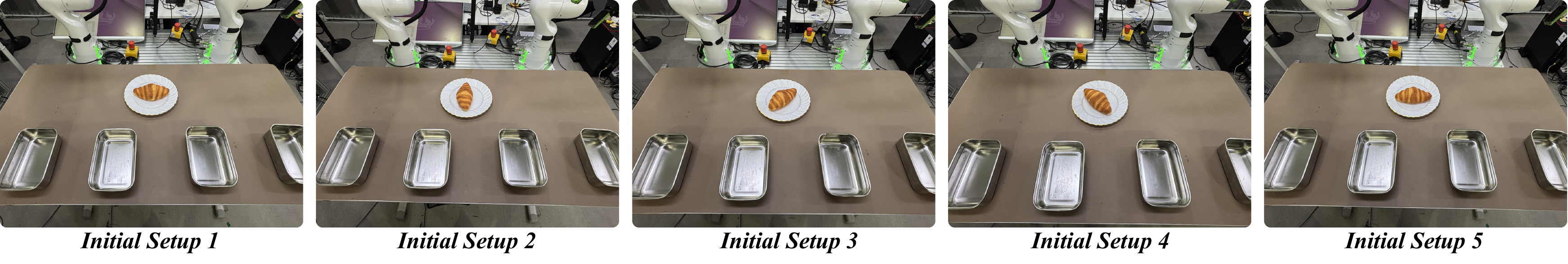}
        \caption{Bread orientations on the plate: horizontal, vertical, anti-diagonal ($/$), diagonal ($\backslash$), and horizontal again.}
        \label{fig:eval-food-storage-bread}
    \end{subfigure}
    \caption{Initial setups for the Food Storage task. (a) The pointing positions define the four setups. (b) The five bread orientations are cycled through within each setup.}
    \label{fig:eval-food-storage}
\end{figure}

\paragraph{Food Storage.}
This task is evaluated over the 4 initial setups shown in Figure~\ref{fig:eval-food-storage-pointing}, each repeated 5 times for 20 trials total. In each trial, the human operator points to one of the four containers, and the robot picks up a piece of bread from the plate and places it in the indicated container. The setups specify which container the human points to, indexed 1 through 4 from the human operator's right. Within each setup, the 5 trials cycle through the bread orientations on the plate shown in Figure~\ref{fig:eval-food-storage-bread}: horizontal, vertical, anti-diagonal ($/$), diagonal ($\backslash$), and horizontal again. For the in-distribution and OOD-silhouette evaluations, the human operator wears pink. For the OOD-clothes evaluation, two clothing colors absent from the training data are used: yellow and green, with 10 trials each.

\begin{figure}[ht]
    \centering
    \includegraphics[width=\linewidth]{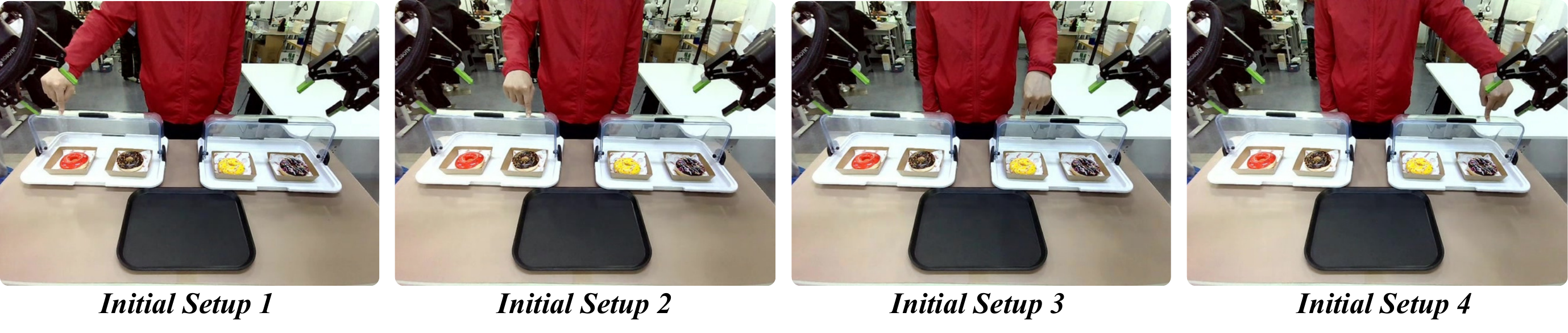}
    \caption{Pointing positions: the donut indexed 1 through 4 from the human operator's right.}
    \label{fig:eval-donut-serving}
\end{figure}
\paragraph{Donut Serving.}
This task is also evaluated over the 4 initial setups shown in Figure~\ref{fig:eval-donut-serving}, each repeated 5 times for 20 trials total. In each trial, the human operator points to one of the four donuts, and the robot picks up a paper to-go box containing pointed donut and places it on the tray. The setups specify which donut the human points to, indexed 1 through 4 from the human operator's right. 5 trials were conducted for each setup. For the in-distribution and OOD-silhouette evaluations, the human operator wears purple. For the OOD-clothes evaluation, two clothing colors absent from the training data are used: orange and red, with 10 trials each.

\subsection{Task workflow}
\label{app:workflow-diagrams}

This subsection presents the workflow structure of each task as a sequence of low-level human and robot subtasks ($H_1, H_2, \ldots$ and $R_1, R_2, \ldots$) connected by precedence edges ($\rightarrow$), following the notation introduced in Section~\ref{sec:task-desgin}. Note that policies are trained on \emph{high-level} task instructions (for example, ``place the bread in the $k$-th container from the left''), not on these low-level subtask sequences. The full set of high-level instructions is released together with the dataset. The low-level breakdowns below are provided to make the workflow structure of each task explicit, particularly the temporal coupling between human and robot actions.

\paragraph{Shelf Cleaning (Collaborator).}~\\

\textit{Human:}
\begin{enumerate}
    \item Hand the Duster to the robot.
    \item Lift the objects on a randomly selected tier of the Shelf.
    \item Once the robot finishes cleaning, lift the objects on the remaining tiers of the Shelf.
    \item Receive the Duster from the robot.
\end{enumerate}

\textit{Robot:}
\begin{enumerate}
    \item Pick up the Duster from the human.
    \item Clean the specific tier of the Shelf with the Duster once objects are removed.
    \item Clean the remaining tier of the Shelf with the Duster once objects are removed.
    \item Hand the Duster back to the human.
\end{enumerate}

\begin{figure}[h]
  \centering
  \includegraphics[width=0.95\linewidth]{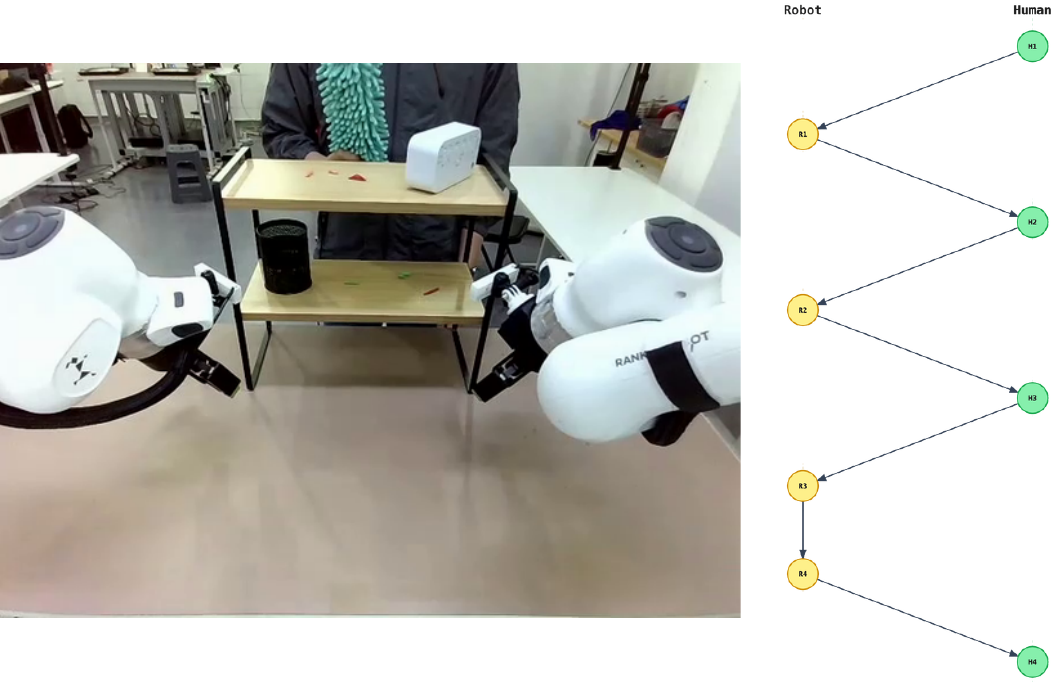}
  \caption{Initial configuration and task workflow for Shelf Cleaning.}
  \label{fig:workflow-shelf-cleaning}
\end{figure}

\paragraph{Table Serving (Collaborator).}~\\

\textit{Human:}
\begin{enumerate}
    \item The human picks up the top napkin from the stack on the human table and walks to the robot table to stand in front of one of the two trays.
    \item When the robot lifts the bowl and the cup, the human unfolds the napkin and lays the napkin flat on the tray.
    \item The human returns to the human table to pick up another napkin and walks to the robot table to stand in front of the tray without a napkin.
    \item When the robot lifts the bowl and the cup, the human unfolds the napkin and lays the napkin flat on the tray.
\end{enumerate}

\textit{Robot:}
\begin{enumerate}
    \item Pick up the Picnic Bowl and Reusable plastic cup from the Handle tray in front of the human's position and hold them in the air.
    \item Place the Picnic Bowl and Reusable plastic cup back onto the Handle tray in front of the human's position.
    \item Pick up the Picnic Bowl and Reusable plastic cup from the Handle tray in front of the human's position and hold them in the air.
    \item Place the Picnic Bowl and Reusable plastic cup back onto the Handle tray in front of the human's position.
\end{enumerate}

\begin{figure}[h]
  \centering
  \includegraphics[width=0.95\linewidth]{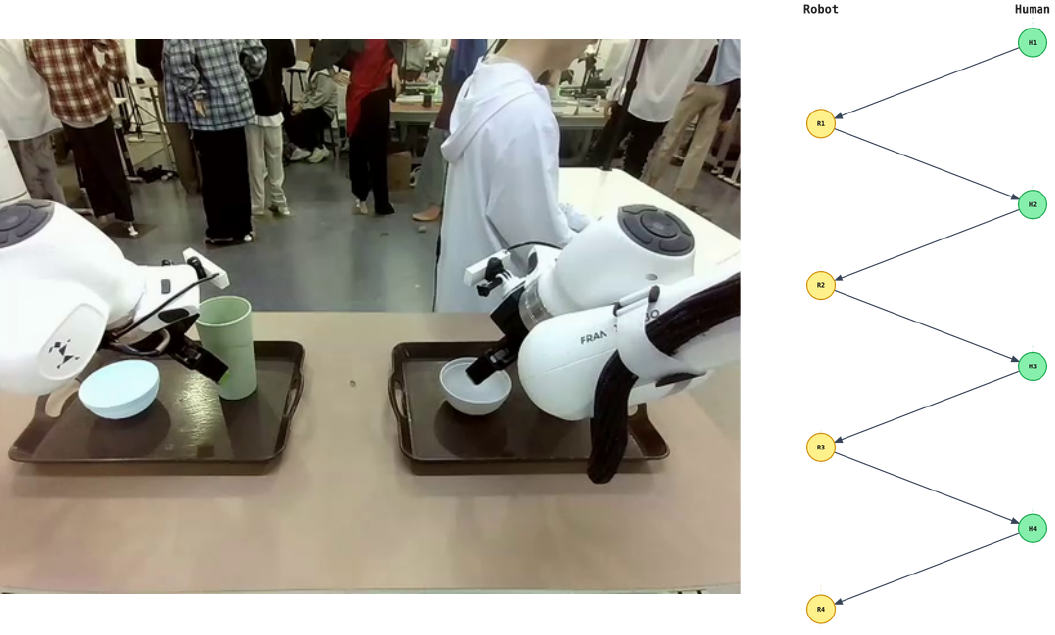}
  \caption{Initial configuration and task workflow for Table Serving.}
  \label{fig:workflow-table-serving}
\end{figure}

\paragraph{Waste Sorting (Coworker).}~\\

\textit{Human:}
\begin{enumerate}
    \item The human picks up one piece of trash that is not a can and places the trash into the appropriate organizing basket.
    \item The human picks up one piece of trash that is not a can and places the trash into the appropriate organizing basket.
    \item The human picks up one piece of trash that is not a can and places the trash into the appropriate organizing basket.
    \item The human picks up one piece of trash that is not a can and places the trash into the appropriate organizing basket.
\end{enumerate}

\textit{Robot:}
\begin{enumerate}
    \item Pick up the can waste from the table and place it in the right Fabric basket.
    \item Pick up the can waste from the table and place it in the right Fabric basket.
\end{enumerate}

\begin{figure}[h]
  \centering
  \includegraphics[width=0.95\linewidth]{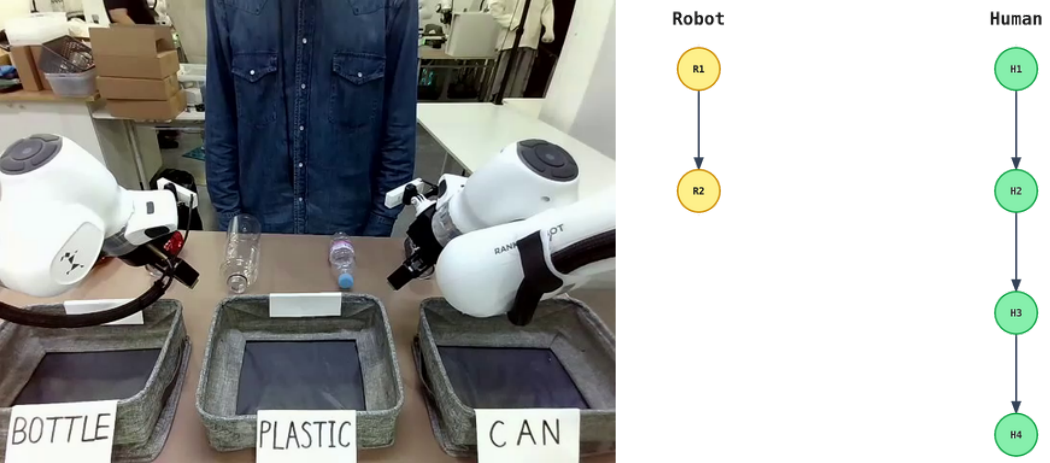}
  \caption{Initial configuration and task workflow for Waste Sorting.}
  \label{fig:workflow-waste-sorting}
\end{figure}

\paragraph{Box Packaging (Coworker).}~\\

\textit{Human:}
\begin{enumerate}
    \item Pick up an object on the table and put it in the box.
    \item Pick up an object on the table and put it in the box.
    \item Pick up an object on the table and put it in the box.
    \item Pick up an object on the table and put it in the box.
    \item Close the lid of the box facing the person.
\end{enumerate}

\textit{Robot:}
\begin{enumerate}
    \item Pick up a Pencil case or Stapler and place it inside the Mailer Box closest to the robot.
    \item Pick up a Pencil case or Stapler and place it inside the Mailer Box closest to the robot.
    \item Close the lid of the Mailer Box closest to the robot.
\end{enumerate}

\begin{figure}[h]
  \centering
  \includegraphics[width=0.95\linewidth]{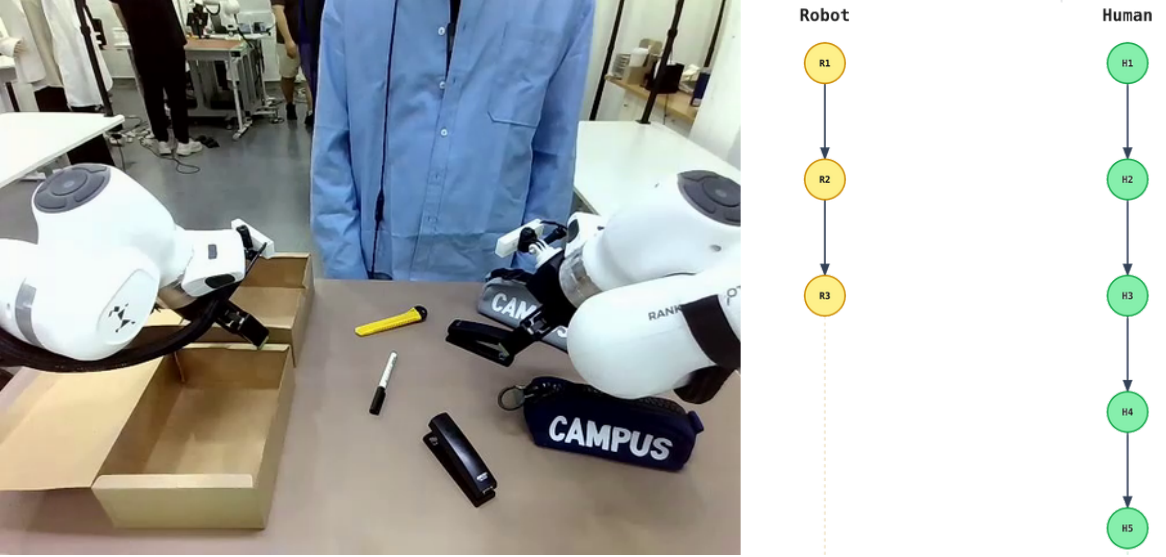}
  \caption{Initial configuration and task workflow for Box Packaging.}
  \label{fig:workflow-box-packaging}
\end{figure}

\paragraph{Food Storage (Supervisor).}~\\

\textit{Human:}
\begin{enumerate}
    \item A person randomly selects and points to an Airtight Container.
\end{enumerate}

\textit{Robot:}
\begin{enumerate}
    \item Place the Butter Roll into the Airtight Container indicated by the human.
\end{enumerate}

\begin{figure}[h]
  \centering
  \includegraphics[width=0.95\linewidth]{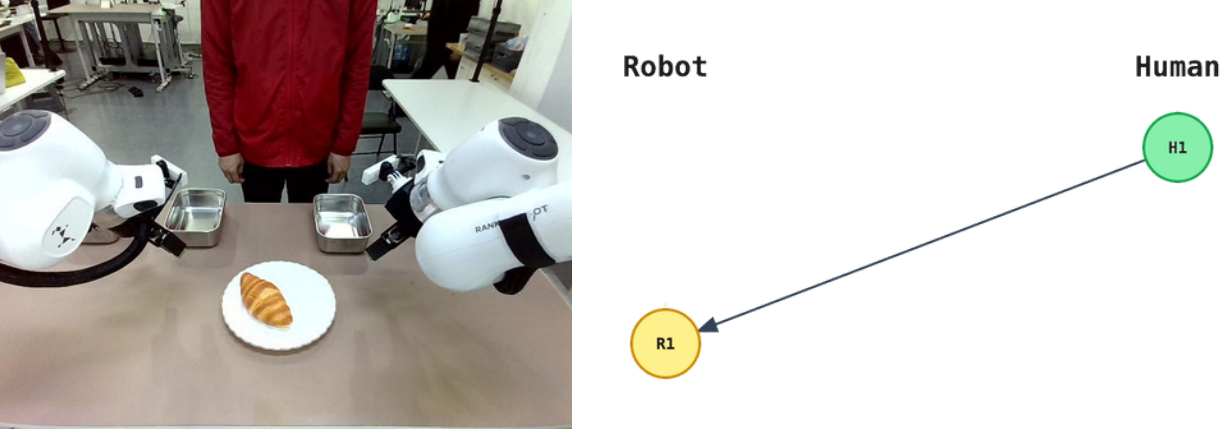}
  \caption{Initial configuration and task workflow for Food Storage.}
  \label{fig:workflow-food-storage}
\end{figure}

\paragraph{Donut Serving (Supervisor).}~\\

\textit{Human:}
\begin{enumerate}
    \item Points to the third donut from the left from the robot's perspective.
\end{enumerate}

\textit{Robot:}
\begin{enumerate}
    \item Pick up the Paper to-go box containing the Donut indicated by the person and place it on the Handle tray.
\end{enumerate}

\begin{figure}[h]
  \centering
  \includegraphics[width=0.95\linewidth]{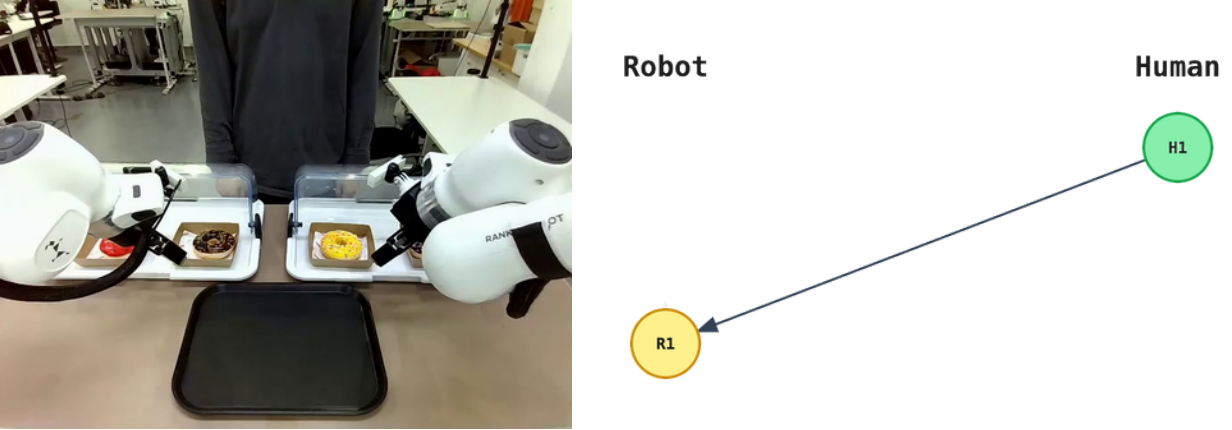}
  \caption{Initial configuration and task workflow for Donut Serving.}
  \label{fig:workflow-donut-serving}
\end{figure}

\section{Model Training Details}
\label{app:model-training}
This section provides the fine-tuning configurations for the two open-source VLAs evaluated throughout the paper, namely $\pi_{0.5}$ and GR00T~N1.6. Within each model, the configuration is held fixed across the Robot-only and HABIT conditions, with dataset being the only factor that differs between conditions. Adjustments specific to the mid-training experiment are described in Appendix~\ref{app:mid-training}. 
All fine-tuning runs are performed on a single node with 8$\times$ H100 GPUs.

\subsection{$\pi_{0.5}$ Fine-Tuning Details}
\label{app:pi05-training}

We fine-tune $\pi_{0.5}$~\citep{pi05}, an open-source VLA model, on our \method~dataset. The model is initialized from the \texttt{pi05\_base} checkpoint and adapted to our bimanual action space. Each action is represented as a $14$-D vector, consisting of a $7$-D Cartesian delta action for each arm, and is zero-padded to the model's architectural $32$-D action dimension. The policy is conditioned on three RGB streams (front, left-wrist, and right-wrist) together with a 14-D proprioceptive state, where each arm contributes a 6-D Cartesian pose and a 1-D gripper state.

For fine-tuning, we use AdamW with gradient clipping at $1.0$, a peak learning rate of $5{\times}10^{-5}$, linear warmup (ratio $0.1$ of training steps) followed by cosine decay to $5{\times}10^{-6}$, and EMA with decay $0.999$. Training is performed in bfloat16, with timestep embeddings and AdamW moment buffers maintained in fp32 for numerical stability. We use $10$-step action chunking horizon. The full hyperparameter configuration is summarized in Table~\ref{tab:pi05-hparams}.

\begin{table}[h]
\centering
\small
\caption{$\pi_{0.5}$ fine-tuning configuration on per-task evaluation}
\label{tab:pi05-hparams}
\begin{tabular}{lll}
\toprule
\textbf{Group} & \textbf{Setting} & \textbf{Value} \\
\midrule
Model & base checkpoint & \texttt{pi05\_base} \\
& action dim (padded) & $32$ \\
& action horizon & $10$ \\
\midrule
Data & images & front, left-wrist, right-wrist (RGB) \\
& state ($14$-D) & per-arm: $3$-D xyz $+$ $3$-D rotation $+$ $1$-D gripper \\
& action ($14$-D) & per-arm: $7$-D Cartesian delta action \\
\midrule
Optimization & optimizer & AdamW \\
& gradient clip & $\lVert g\rVert\!\le\!1.0$ \\
& peak learning rate & $5{\times}10^{-5}$ \\
& schedule & linear warmup $\rightarrow$ cosine decay \\
& warmup ratio & $0.1$ ($\times$ training steps) \\
& end LR (cosine) & $5{\times}10^{-6}$ \\
& training steps & $5{,}000$ (mid-training adjustments in Appendix~\ref{app:mid-training}) \\
& batch size & $128$ (fine-tuning) / 256 (mid-training) \\
& EMA decay & $0.999$ \\
& precision & bfloat16 \\
& seed & $42$ \\
\bottomrule
\end{tabular}
\end{table}

\subsection{GR00T N1.6 Fine-Tuning Details}
\label{app:groot-training}

We fine-tune NVIDIA's GR00T~N1.6~\citep{groot} as a second VLA baseline on our
HABIT dataset. The model is initialized from the open-weights
\texttt{GR00T-N1.6-3B} checkpoint, which couples an internal NVIDIA Cosmos-2B VLM variant as a backbone with a diffusion-based action decoder. To preserve the pretrained
representations under our limited per-task data budget, we tune only the top $4$ transformer layers of the LLM backbone, the multimodal projector, and the diffusion action decoder, keeping the visual encoder and the remaining LLM layers frozen. The model is adapted to our bimanual action space using a $14$-D state and $14$-D action vector. The state vector concatenates, per arm, a $3$-D Cartesian xyz position, a $3$-D rotation, and a $1$-D gripper signal. Each action is represented as a $14$-D vector, consisting of a $7$-D Cartesian delta action for each arm. The policy is conditioned on three RGB streams (front, left-wrist, and right-wrist).

For fine-tuning, we use AdamW with gradient
clipping at $1.0$, a peak learning rate of $1{\times}10^{-4}$, weight decay
$1{\times}10^{-5}$, linear warmup (ratio $0.05$) followed by cosine decay, and
color-jitter augmentation on input frames. Training is performed in bfloat16
with a $16$-step action chunking horizon. The full hyperparameter
configuration is summarized in Table~\ref{tab:groot-hparams}.

\begin{table}[h]
\centering
\small
\caption{GR00T~N1.6 fine-tuning configuration.}
\label{tab:groot-hparams}
\begin{tabular}{lll}
\toprule
\textbf{Group} & \textbf{Setting} & \textbf{Value} \\
\midrule
Model & base checkpoint & \texttt{nvidia/GR00T-N1.6-3B} \\
& backbone & Eagle-Block2A-2B-v2 (frozen) \\
& visual encoder & frozen \\
& tuned modules & top-4 LLM layers $+$ multimodal projector $+$ diffusion action head \\
& action horizon & $16$ \\
\midrule
Data & images & front, left-wrist, right-wrist (RGB) \\
& state ($14$-D) & per-arm: $3$-D xyz $+$ $3$-D rotation $+$ $1$-D gripper \\
& action ($14$-D) & per-arm: $7$-D Cartesian delta action \\
& augmentation & color jitter (br.\ $0.3$, cont.\ $0.4$, sat.\ $0.5$, hue $0.08$) \\
\midrule
Optimization & optimizer & AdamW \\
& gradient clip & $\lVert g\rVert\!\le\!1.0$ \\
& peak learning rate & $1{\times}10^{-4}$ \\
& weight decay & $1{\times}10^{-5}$ \\
& schedule & linear warmup $\rightarrow$ cosine decay \\
& warmup ratio & $0.05$ \\
& training steps & $5{,}000$ (mid-training adjustments in Appendix~\ref{app:mid-training}) \\
& batch size & $128$ \\
& precision & bfloat16 \\
& seed & default \\
\bottomrule
\end{tabular}
\end{table}

\section{Main Experiment Details}
\label{app:main-experiment}
This section provides the experiment-specific details for the main experiment (Section~\ref{sec:exp-main}), including per-task training data statistics, the success criteria used during evaluation, and failure analysis for the remaining tasks not covered in the main text. Model architecture and fine-tuning hyperparameters shared across the experiments are described in Appendix~\ref{app:model-training}.

\subsection{Training Data Statistics}
\label{app:training-data}

Table~\ref{tab:training-data} summarizes the data used for fine-tuning. HABIT episodes are on average longer than Robot-only ones because human-robot interaction takes time. Supervisor tasks are disproportionately longer because the human's wait time before pointing is deliberately varied during data collection (Appendix~\ref{app:task-setup}), preventing policies from learning a short-cut response to the language instruction alone and forcing them to ground the gesture before acting.

\begin{table}[h]
  \caption{Training data per task. Robot-only counts episodes collected without a co-present human (where applicable), while \method~counts episodes collected with a co-present human.}
  \label{tab:training-data}
  \centering
  \small
  \begin{tabular}{llrrrr}
    \toprule
    & & \multicolumn{2}{c}{Robot-only} & \multicolumn{2}{c}{HABIT} \\
    \cmidrule(lr){3-4} \cmidrule(lr){5-6}
    Role & Task & \#Episodes & Hours & \#Episodes & Hours \\
    \midrule
    Collaborator & Table Serving   & N/A & N/A  & 206 & 3.34 \\
                 & Shelf Cleaning  & N/A & N/A  & 207 & 2.34 \\
    \midrule
    Coworker     & Waste Sorting   & 202 & 1.58 & 189 & 1.65 \\
                 & Box Packaging   & 212 & 2.66 & 203 & 2.73 \\
    \midrule
    Supervisor   & Food Storage    & 211 & 1.04 & 203 & 1.82 \\
                 & Donut Serving   & 204 & 0.94 & 204 & 1.66 \\
    \bottomrule
  \end{tabular}
\end{table}

\subsection{Evaluation Details}
\label{app:rollout-protocol}
For each (task, model, condition) cell, we run $N = 20$ independent trials following the predefined task-specific evaluation protocol described in Appendix~\ref{app:eval-protocol}. The robot's initial configuration is held fixed across trials, and object placements are reset to their data-collection positions with only minor positional noise. Our goal is to evaluate human-aware behavior rather than manipulation robustness to object placement, so we fix the initial state to isolate this signal. The same human operator executes all in-distribution trials, wearing the most-frequently-recorded clothing color for that task. OOD evaluation conditions are described in Appendix~\ref{app:ood}. All trials are scored using the success rate defined in Eq.~\ref{eq:hsr}.

\paragraph{Inference settings.}
Although the models are trained with action horizons of 10 for $\pi_{0.5}$ and 16 for GR00T N1.6, only the first 4 steps of each predicted action chunk are executed before the next query, with the remaining steps discarded. This ensures responsive control and consistent execution frequency across the two models.

\subsection{Failure Analysis for Remaining Tasks}
\label{app:failure-analysis-extra}
\begin{figure}[h]
  \centering
  \includegraphics[width=1.0\linewidth]{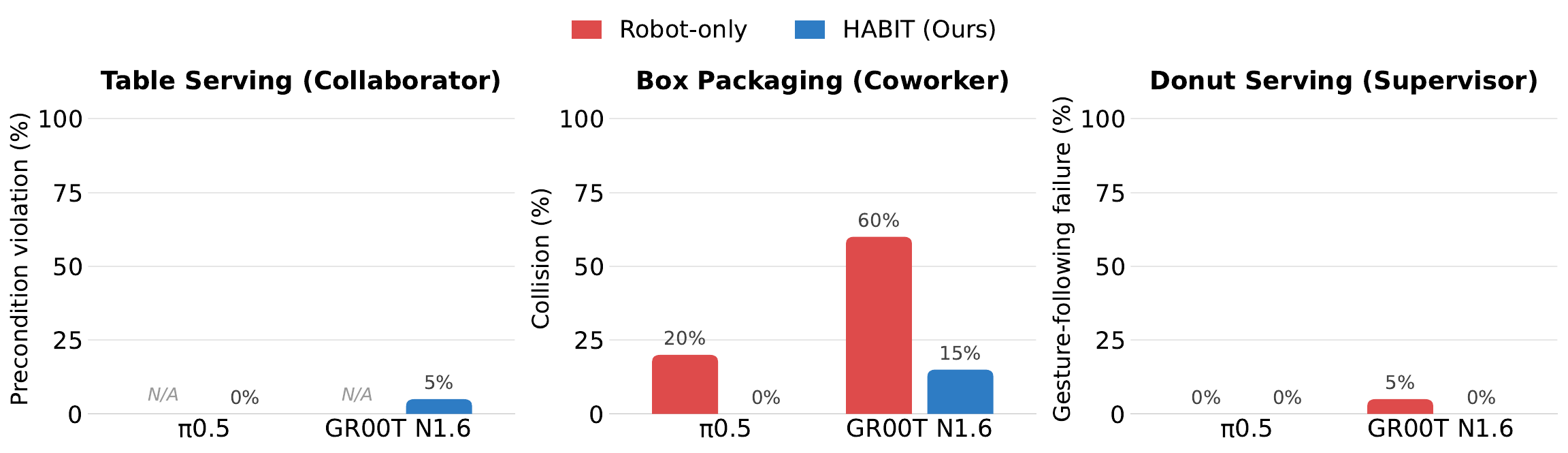}
  \caption{Role-specific failure analysis on the remaining three tasks. Failure types are equal to those in Figure~\ref{fig:failure-analysis}. Robot-only is not applicable for the Collaborator task.}
  \label{fig:failure-analysis-extra}
\end{figure}

Section~\ref{sec:failure} presents detailed failure analysis for one representative task per role, namely Shelf Cleaning (Collaborator), Waste Sorting (Coworker), and Food Storage (Supervisor). Figure~\ref{fig:failure-analysis-extra} reports the same failure mode breakdown for the remaining three tasks, namely Table Serving (Collaborator), Box Packaging (Coworker), and Donut Serving (Supervisor). The patterns are consistent with the representative tasks. On Box Packaging, HABIT-trained policies sharply reduce collisions relative to Robot-only baselines for both $\pi_{0.5}$ and GR00T N1.6. On Donut Serving, the two regimes match because the indexed language instruction alone is sufficient to identify the target unambiguously, as discussed in Section~\ref{sec:exp-main}. On Collaborator Table Serving, $\pi_{0.5}$ achieves a near-perfect success rate with HABIT training while GR00T's manipulation instability dominates the failure breakdown.

\subsection{Event-Level Evaluation}
\label{app:event-metrics}
Beyond task success rates and the role-specific failure analysis in Section~\ref{sec:experiments}, we measure one event-level metric per role to characterize human-aware behavior in finer detail. We manually annotate the videos of the in-distribution trials in Section~\ref{sec:exp-main} for the \method-trained policies.
\begin{itemize}[leftmargin=1.5em, itemsep=1mm, topsep=2pt]
\item \textit{Synchronization lag.} For Collaborator tasks, the time from the human completing their subtask to the robot reaching the target of its next subtask.
\item \textit{Yield rate.} For Coworker tasks, the fraction of conflict events in which the robot visibly retreats instead of colliding. A conflict event occurs when the human's arm enters the robot's workspace while the robot is acting.
\item \textit{Gesture response time.} For Supervisor tasks, the time from gesture onset to the robot reaching its next target, which is the indicated donut box in Donut Serving and the bread in Food Storage.
\end{itemize}
VLA policies continuously produce small motions, so the exact onset of a robot motion is hard to determine. We therefore stop timing when the gripper arrives at the target, just before grasping or acting on it. We count only events completed before any failure in a trial. Gesture response time is averaged over trials in which the policy follows the gesture and completes the task, since whether the policy acts on the indicated target is already captured by the gesture-following failure in Section~\ref{sec:failure}.

Table~\ref{tab:event-metrics} reports the results. $\pi_{0.5}$ synchronizes with the human faster, yields in every conflict event, and responds to gestures faster than GR00T N1.6. GR00T N1.6 also shows larger variance in timing. This ordering is consistent with the success rates in Figure~\ref{fig:main-results}, suggesting that the event-level metrics reflect the quality of human-aware behavior beyond task success.

\begin{table}[h]
\centering
\small
\caption{Event-level metrics of \method-trained policies. Sync. lag denotes synchronization lag. Timings are reported as mean $\pm$ standard deviation in seconds, and $n$ denotes the number of events.}
\label{tab:event-metrics}
\begin{tabular}{l l l c c c c}
\toprule
& & & \multicolumn{2}{c}{$\pi_{0.5}$} & \multicolumn{2}{c}{GR00T N1.6} \\
\cmidrule(lr){4-5}\cmidrule(lr){6-7}
Role & Metric & Task & Value & $n$ & Value & $n$ \\
\midrule
Collaborator & Sync. lag $\downarrow$ & Shelf Cleaning & 2.47 $\pm$ 0.90 & 52 & 4.43 $\pm$ 3.19 & 31 \\
             &                       & Table Serving  & 4.61 $\pm$ 1.80 & 79 & 7.56 $\pm$ 2.93 & 31 \\
Coworker     & Yield rate $\uparrow$ & Waste Sorting  & 1.00 & 43 & 0.89 & 47 \\
             &                       & Box Packaging  & 1.00 & 49 & 0.95 & 63 \\
Supervisor   & Response time $\downarrow$ & Food Storage  & 3.02 $\pm$ 0.62 & 15 & 6.37 $\pm$ 4.39 & 5 \\
             &                       & Donut Serving  & 7.07 $\pm$ 4.65 & 19 & 9.89 $\pm$ 4.75 & 13 \\
\bottomrule
\end{tabular}
\end{table}

\subsection{Training on Anonymized Data}
\label{app:anonymization}
The results in the main paper were obtained with policies trained on data before face anonymization. To verify that the released anonymized data supports the same conclusions, we retrain $\pi_{0.5}$ on face-anonymized versions of the same training sets for all six evaluation tasks, and evaluate each policy under the same protocol over 20 trials. As shown in Table~\ref{tab:anonymization}, success rates are comparable on every task, differing by at most a single trial, and the role-specific failure rates are identical between the two conditions. This indicates that facial appearance is not a critical cue for the behaviors targeted by \method, as the policy relies on body posture, arm motion, and workspace context, all of which are preserved under face blurring.

\begin{table}[h]
\centering
\small
\caption{Success rates of $\pi_{0.5}$ trained on original and face-anonymized data. Success rates are reported in percent.}
\label{tab:anonymization}
\begin{tabular}{l l c c}
\toprule
Task & Role & Original & Anonymized \\
\midrule
Table Serving & Collaborator & 95.0 & 95.0 \\
Shelf Cleaning & Collaborator & 45.0 & 50.0 \\
Waste Sorting & Coworker & 65.0 & 70.0 \\
Box Packaging & Coworker & 85.0 & 90.0 \\
Donut Serving & Supervisor & 95.0 & 95.0 \\
Food Storage & Supervisor & 70.0 & 75.0 \\
\bottomrule
\end{tabular}
\end{table}

\section{OOD Robustness Analysis}
\label{app:ood}

\subsection{Motivation}

In prior work on robot manipulation, out-of-distribution (OOD) evaluation has typically focused on axes such as object placement, distractors, and lighting~\citep{droid, walke2023bridgedata}. For human-robot interaction, the most consequential distribution shift at deployment time is the human itself. A deployed robot will encounter people whose clothing and body silhouette differ from those of the data collectors. We evaluate \method-trained policies along these axes directly.

\subsection{Evaluation Conditions}

We construct three evaluation cells per task, applied to both \method-trained $\pi_{0.5}$ and GR00T~N1.6. Each cell contains 20 trials, using the same evaluation protocol as the main experiment.

\begin{itemize}
    \item \textbf{In-distribution}: The original human operator wears the most-frequently-recorded clothing color for the task. Identical to the corresponding cell in Section~\ref{sec:exp-main}.
    \item \textbf{OOD-clothing}: The original human operator wears two clothing colors that were not seen during training (10 trials per color). The clothing rotation in our collection protocol (Section~\ref{sec:data-collection}) ensures that the held-out colors are genuinely unseen rather than rare-but-present.
    \item \textbf{OOD-silhouette}: Two human operators not present in the training data execute the trials (10 trials per operator). Three operators with distinct body silhouettes participate in the evaluation: slim (162 cm / 57 kg), athletic (180 cm / 77 kg), and heavier (170 cm / 120 kg). The operator seen during training is the slim silhouette for Table Serving and the athletic silhouette for the remaining five tasks; the other two silhouettes are treated as OOD.
\end{itemize}

\subsection{Per-Task Results}

\begin{figure}[t]
  \centering
  \begin{subfigure}{\linewidth}
    \centering
    \includegraphics[width=1.0\linewidth]{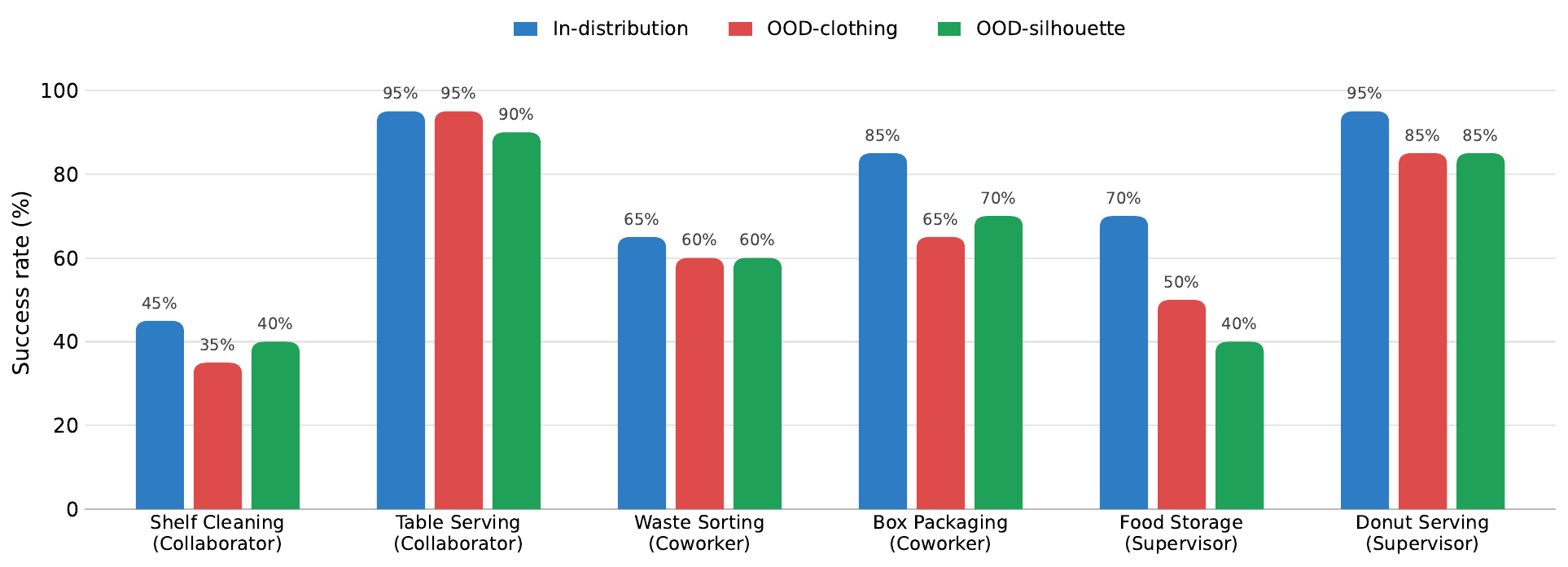}
    \caption{$\pi_{0.5}$}
    \label{fig:ood-results-pi05}
  \end{subfigure}
  \begin{subfigure}{\linewidth}
    \centering
    \includegraphics[width=1.0\linewidth]{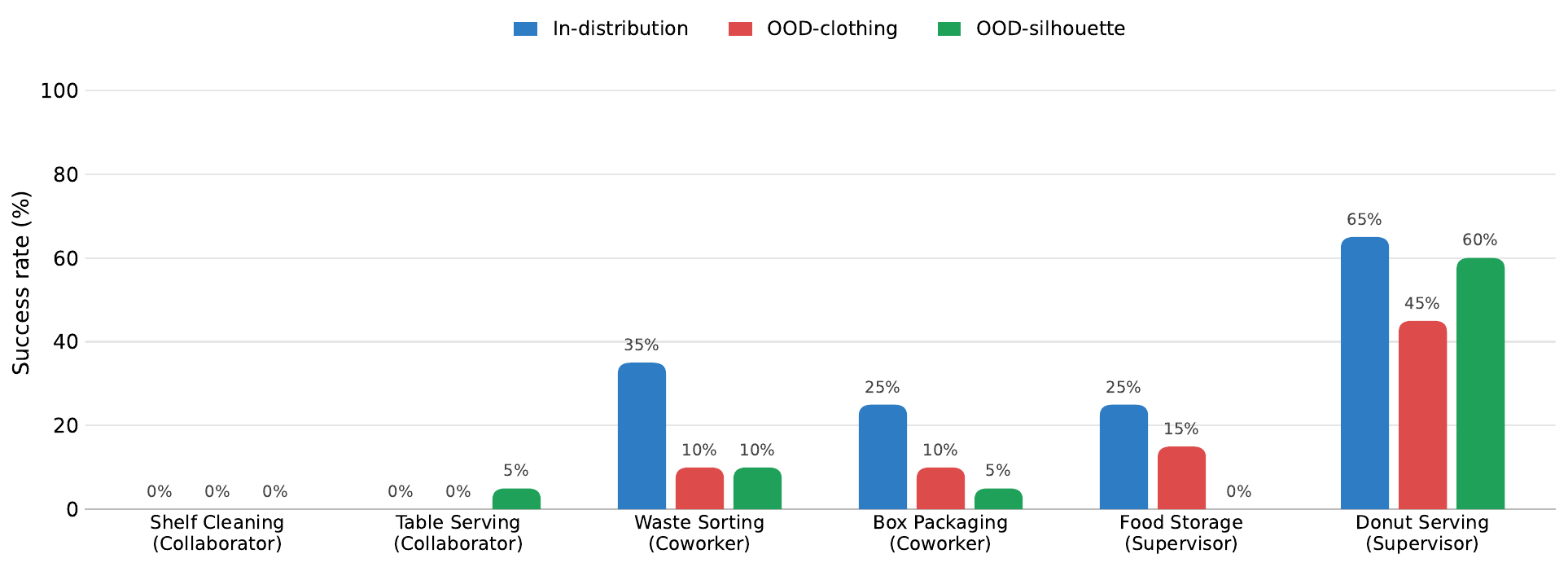}
    \caption{GR00T N1.6}
    \label{fig:ood-results-groot}
  \end{subfigure}
  \caption{Success rate under in-distribution and out-of-distribution conditions for HABIT-trained (a) $\pi_{0.5}$ and (b) GR00T N1.6. Each cell reports the success rate over 20 trials. In-distribution column is identical to the corresponding cell in Figure~\ref{fig:main-results}.}
  \label{fig:ood-hsr}
\end{figure}

\begin{figure}[h]
  \centering
  \includegraphics[width=1.0\linewidth]{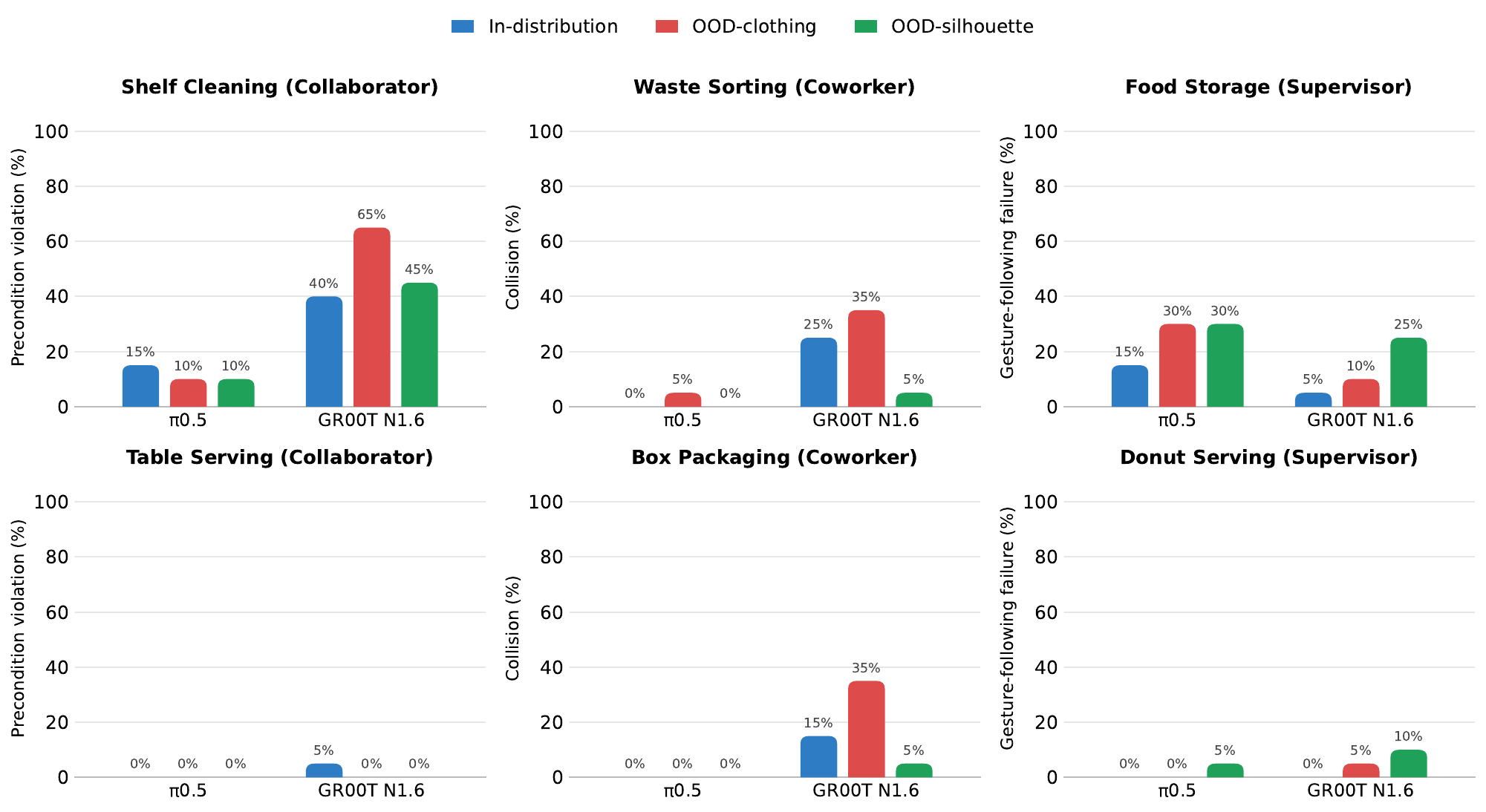}
  \caption{Role-specific failure analysis under in-distribution and out-of-distribution conditions. Failure types are equal to those in Figure~\ref{fig:failure-analysis}.}
  \label{fig:ood-failure-analysis}
\end{figure}

Figure~\ref{fig:ood-hsr} reports success rates across the three evaluation conditions for both $\pi_{0.5}$ and GR00T N1.6. Figure~\ref{fig:ood-failure-analysis} further decomposes results into role-specific failure modes. Overall, human-centric distribution shifts lead to modest performance degradation. Across both models, performance degradation is primarily driven by manipulation failures, while role-specific behaviors such as reactive yielding on Coworker tasks and spatiotemporal synchronization on Collaborator tasks remain largely stable under OOD conditions. The main exception is the Supervisor tasks, where OOD-silhouette increases gesture-following failures, suggesting that grounding pointing gestures is more sensitive to changes in human appearance. We hypothesize that the diversity intentionally introduced during data collection, including rotating clothing colors and multiple human operators, contributes to robustness against human-centric distribution shifts.

\subsection{Additional Generalization Evaluations}
\label{app:ood-extra}
Beyond human-centric shifts in clothing and body silhouette, we evaluate the \method-trained $\pi_{0.5}$ policies under two additional distribution shifts, using the same checkpoints without any further fine-tuning, with 20 trials per condition.

\paragraph{Unseen background.} We enclose the workspace with a nature-scene backdrop that replaces the lab background in all camera views. Success rates remain comparable to the original setting, at 65\% versus 65\% on Waste Sorting and 65\% versus 70\% on Food Storage. All failures are manipulation errors, and no collisions occur on Waste Sorting.

\paragraph{Multiple humans.} We evaluate Waste Sorting with two humans standing side by side and alternating turns, so that their movements naturally overlap at times. The success rate matches the single-human setting at 65\%, and the policy avoids collisions in 44 of 46 conflict events. Both collisions occur when the two humans move simultaneously, a conflict pattern not covered by the one-human, one-robot configuration of our data collection.

\paragraph{Summary and scope.} Table~\ref{tab:ood-summary} summarizes the out-of-distribution performance of the \method-trained $\pi_{0.5}$ policy. Across all shifts, performance degrades only modestly, and human-aware behaviors such as yielding and coordination are largely preserved. We note, however, that the unseen-background evaluation replaces the background while keeping the same tabletop, so it bounds sensitivity to background appearance rather than demonstrating transfer to an entirely new environment. Likewise, simultaneous movement of multiple humans remains outside the scope of the current dataset.

\begin{table}[h]
\centering
\small
\caption{Summary of out-of-distribution performance of the HABIT-trained $\pi_{0.5}$ policy. Success rates are reported in percent.}
\label{tab:ood-summary}
\begin{tabular}{l l c c}
\toprule
Distribution shift & Scope & In-distribution & Out-of-distribution \\
\midrule
Clothing color & 6 tasks, average & 75.8 & 65.0 \\
Body silhouette & 6 tasks, average & 75.8 & 64.2 \\
Unseen background & Waste Sorting & 65.0 & 65.0 \\
Unseen background & Food Storage & 70.0 & 65.0 \\
Multiple humans & Waste Sorting & 65.0 & 65.0 \\
\bottomrule
\end{tabular}
\end{table}

\section{Mid-Training Experiment Details}
\label{app:mid-training}

This section provides additional details for the mid-training experiment in Section~\ref{sec:exp-scaling}, including the mid-training subset construction and the hyperparameter adjustments specific to this experiment.

\subsection{Mid-Training Subset}
\label{app:mid-training-subset}

We construct the mid-training subset from HABIT by sampling up to 100 demonstrations per task from 41 tasks, with the 6 evaluation tasks (Section~\ref{sec:evaluation}) excluded. The 41 tasks span all three roles (Collaborator, Coworker, Supervisor). For tasks with fewer than 100 collected demonstrations, all available demonstrations are included.

\subsection{Hyperparameter Adjustments}
\label{app:mid-training-hyperparams}

Mid-training is performed on $\pi_{0.5}$ only, starting from the same \texttt{pi05\_base} initialization as the main experiment. The mid-training stage uses the same architecture and most hyperparameters in the main experiment (Appendix~\ref{app:pi05-training}), with the following adjustments to accommodate the larger dataset.

\begin{itemize}[leftmargin=1.5em, itemsep=1mm, topsep=1pt]
    \item Batch size is increased from 128 to 256.
    \item The total number of training steps is set to 16{,}500, corresponding to 2 epochs over the mid-training subset.
    \item The warmup phase is set to 1{,}650 steps, preserving the warmup ratio of 0.1.
\end{itemize}

All other hyperparameters (learning rate schedule, EMA, optimizer settings, and action horizon) are held identical to the main experiment fine-tuning configuration.

\paragraph{Fine-tuning stage.}
After mid-training, fine-tuning on each evaluation task uses the same hyperparameters as the main experiment, with two exceptions that scale with the demonstration count, namely the total training steps and the warmup steps. Both quantities are scaled proportionally to the number of demonstrations, namely 1{,}250 steps with 125 warmup steps for 50 demonstrations, 2{,}500 steps with 250 warmup steps for 100 demonstrations, and 5{,}000 steps with 500 warmup steps for 200 demonstrations. The warmup ratio of 0.1 is preserved across all settings. Direct fine-tuning baselines use the same scaled hyperparameters but skip the mid-training stage, starting directly from \texttt{pi05\_base}.

\subsection{Failure Analysis}
\label{app:mid-training-failure}

\begin{figure}[h]
  \centering
  \begin{subfigure}{\linewidth}
    \centering
    \includegraphics[width=1.0\linewidth]{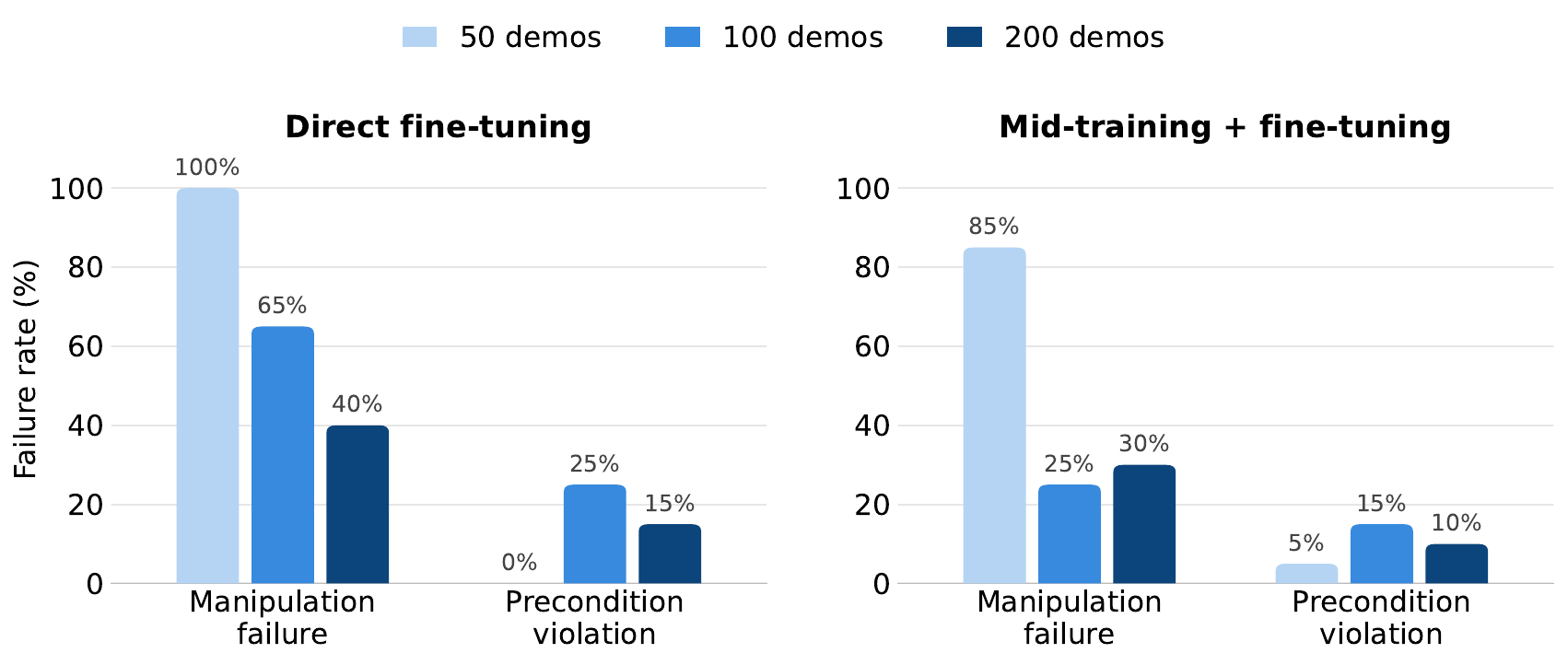}
    \caption{Failure analysis on Shelf Cleaning task}
    \label{fig:mid-training-failure-shelf}
  \end{subfigure}
  \begin{subfigure}{\linewidth}
    \centering
    \includegraphics[width=1.0\linewidth]{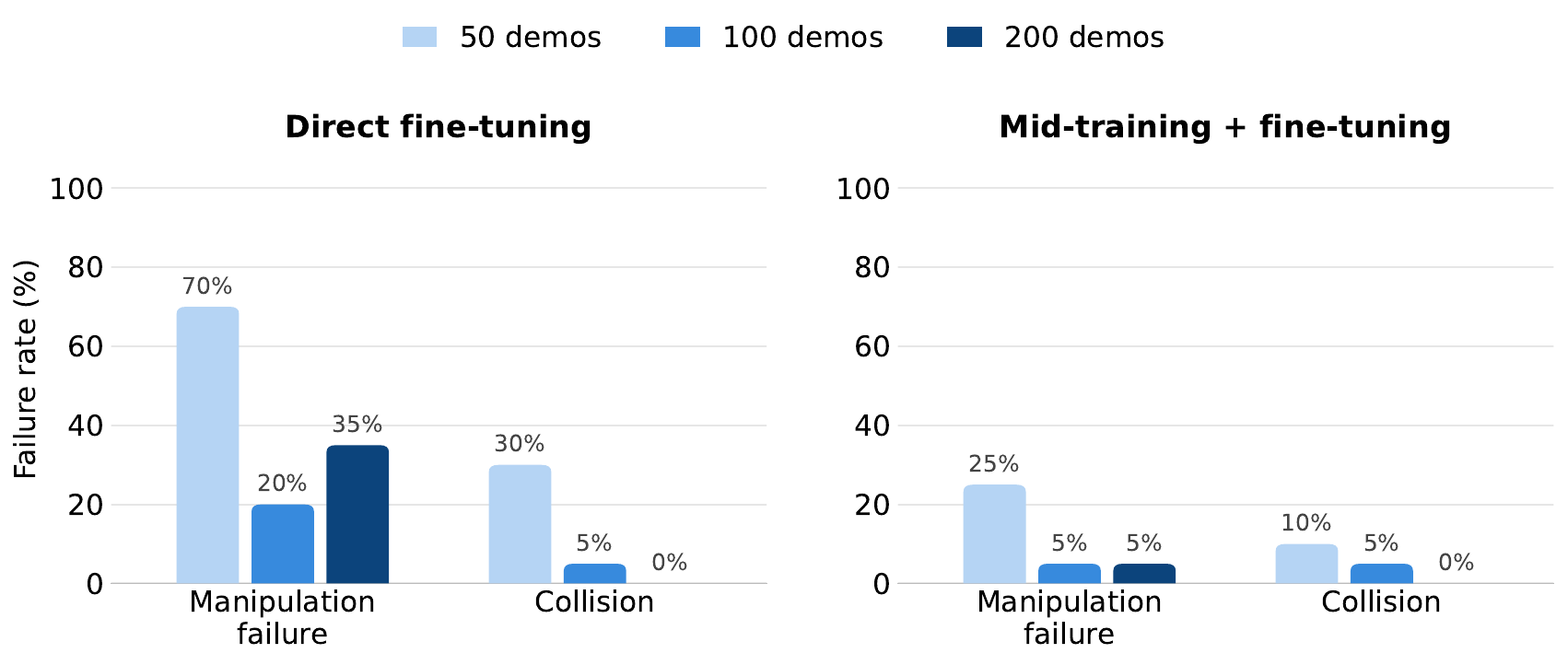}
    \caption{Failure analysis on Waste Sorting task}
    \label{fig:mid-training-failure-waste}
  \end{subfigure}
  \caption{Failure mode breakdown for direct fine-tuning and HABIT mid-training followed by fine-tuning on two unseen human-robot interaction tasks. Results are shown for $\pi_{0.5}$ using 50, 100, and 200 task-specific demonstrations. Each bar reports the fraction of trials attributed to manipulation failures or the task-specific failure modes.}
  \label{fig:mid-training-failure}
\end{figure}

Figure~\ref{fig:mid-training-failure} reports the failure mode breakdown for direct fine-tuning and HABIT mid-training followed by fine-tuning across the three demonstration counts, illustrating how HABIT mid-training improves adaptation to new human-robot interaction tasks. With only 50 or 100 task-specific demonstrations, direct fine-tuning is still dominated by manipulation failures, preventing the policy from reliably executing the interaction behaviors required by the task. In contrast, mid-trained policies exhibit substantially fewer manipulation failures while already maintaining low rates of role-specific failures on both Collaborator and Coworker tasks. As the number of demonstrations increases, the gap narrows, but mid-training consistently achieves lower failure rates. These results suggest that HABIT mid-training provides a strong prior that allows policies to rapidly acquire new human-robot interaction tasks from limited demonstrations.

\section{Combining \method~with Human-Absent Data}
\label{app:cotrain}

We examine whether using \method~with existing human-absent robot demonstration data benefits the policy. To test this, we co-train \method~with Robot-only and evaluate the resulting policy in settings both with and without a human. On Waste Sorting and Food Storage, we compare three $\pi_{0.5}$ policies trained on Robot-only, \method, and their union. All three share the same training configuration as in Section~\ref{sec:exp-setup}, so the co-trained policy sees no more gradient updates than the single-source conditions. Each policy is evaluated over 20 trials each with and without a human. Without a human, the robot clears only the cans in Waste Sorting and follows the indexed language instruction alone in Food Storage.

\begin{figure}[t]
  \centering
  \includegraphics[width=\linewidth]{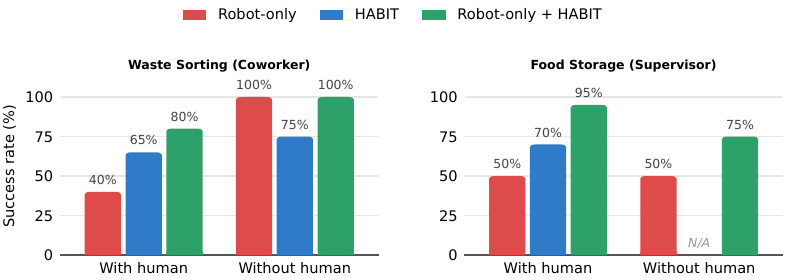}
  \caption{Success rates of $\pi_{0.5}$ trained on Robot-only, \method, and Robot-only + \method, evaluated with and without a co-present human. Co-training on Robot-only and \method~achieves the highest success rate in all four settings. N/A: without a human, the policy trained on \method~alone waits for a gesture that never arrives, so success rate is not measurable.}
  \label{fig:cotrain}
\end{figure}

As shown in Figure~\ref{fig:cotrain}, without a human, the policy trained on \method~alone shows a moderate drop on Waste Sorting but still completes most trials. On Food Storage, it is trained to act on the human's pointing gesture, so without a human it waits for a cue that never arrives. Co-training resolves both cases, recovering or improving performance in the without-human setting while also outperforming both single-source policies when a human is present. We hypothesize that robot-only demonstrations strengthen core manipulation skills while \method~contributes human-aware behaviors on top of them. These results suggest that human presence is an important axis of data diversity that complements, rather than replaces, existing human-absent data.

\section{Ethics Statement}
\label{app:ethics}

\method~was constructed under a human-subjects research protocol following established principles for ethical human-subjects research. Because every episode of \method~contains a co-present human partner whose body, clothing, and gestures are recorded in identifiable form, we describe below the consent procedures, internal ethics review, risk-mitigation measures, privacy protections, and participant rights that govern this dataset. The signed consent form, the data-protection plan, and the internal ethics-review record are available from the authors upon request.

\paragraph{Participant consent.}
All human collectors who appear in \method~are full-time employees of the authors' institution, recruited and trained as data-collection staff. Every participant signed a written consent form (version~1.0, December~2025) before any data containing them was retained. The consent form discloses several items. First, the dataset is released publicly under the CC~BY~4.0 license through Hugging~Face Datasets and is therefore freely redistributable by third parties. Second, the five-camera setup and the precise data items recorded are described, namely RGB video, robot kinematics, and metadata, with no audio collected at any stage. Third, the physical, ergonomic, and identifiability risks are explained together with the corresponding mitigations. Fourth, the participant retains the right to refuse or withdraw at any time without consequence. Finally, the procedural scope and limitations of data takedown after public release are explained.
Participation is documented as voluntary, and refusal or withdrawal is explicitly guaranteed not to affect employment or performance evaluations.
 
\paragraph{Compensation.}
Because participation occurs within scheduled work hours under existing full-time employment contracts, compensation takes the form of regular salary rather than a separate research stipend. The applicable wage exceeds the local statutory minimum wage. No additional fees are charged to participants.

\paragraph{Internal ethics review.}
Prior to the start of data collection, we conducted a documented internal ethics review. The review involved three reviewers occupying \emph{distinct} roles, namely the principal investigator (submitter), an independent ethics reviewer drawn from senior leadership outside the data-collection team, and a data-security reviewer. The review produced a signed record covering six risk categories: physical safety, physical discomfort, hygiene, identifiable-data exposure, voluntariness/coercion, and data security. The record verifies the consent procedure, the right-to-withdraw infrastructure, and the data-protection plan summarized below.

\paragraph{Risk identification and mitigation.}
The internal review identified and mitigated risks across six categories.
\emph{Physical safety} risks from operating around a bimanual robot are mitigated by an always-available emergency-stop button, mandatory pre-session safety briefings on safe-distance operation, and dress-code restrictions on loose accessories.
\emph{Physical discomfort}, particularly from the head-mounted egocentric camera, is mitigated by a 45-minute work / 15-minute rest cycle and an explicit, written right to pause or stop at any time.
\emph{Hygiene} risks from shared equipment are mitigated by alcohol-swab sterilization between users.
\emph{Identifiable-data exposure} is addressed through the privacy protections and takedown procedures described below.
\emph{Coercion} risk is mitigated by the written guarantee that refusal does not affect employment evaluations and by the availability of an independent escalation channel to the ethics reviewer outside the participant's reporting line.
\emph{Data security} is addressed by the data-protection plan.

\paragraph{Privacy protection.}
While the signed consent form authorized release of fully identifiable video on the grounds that body silhouette, clothing, and gesture function as the dataset's principal learning signals, we elected to apply an \textbf{additional layer of privacy protection beyond what consent required}: faces of all human participants are blurred in the public release. Body silhouette, clothing, posture, and gestural cues, which constitute the actual signals analyzed in this work (e.g.\ the OOD evaluations on clothing color and body silhouette in Appendix~\ref{app:ood}), are preserved. No audio is recorded at any stage of the pipeline. Direct identifiers (names, contact information, signed consent forms) are stored in a repository physically and logically isolated from the released dataset, with access restricted to the principal investigator alone via a separate IAM role.

\paragraph{Data security.}
The released dataset and the underlying raw recordings are stored on Amazon~S3 with TLS in transit and AES-256 server-side encryption at rest. All access is governed by AWS~IAM under a least-privilege policy and protected by mandatory multi-factor authentication. Access events are logged via AWS~CloudTrail. The storage provider holds ISO/IEC~27001, 27017, 27018, and SOC~2 certifications. Public-access blocks are enabled on all buckets, and a quarterly access-permission review is conducted with immediate revocation upon role change.

\paragraph{Right to withdraw.}
Participants retain the right to request withdrawal of their data at any time, and the consent form establishes a dedicated takedown email channel for this purpose.
\emph{Before} public release, a withdrawal request results in deletion of all video and derived data (including backups and any local working copies) containing the participant, and removal from internal training and evaluation pipelines.
\emph{After} public release, a takedown request results in three actions. The participant's episodes are removed from all author-controlled mirrors (including the Hugging~Face release) within fourteen business days of receipt. These episodes are then permanently excluded from all future redistributions. They are also removed from internal training and evaluation pipelines.
The consent form is explicit, and we wish to repeat here, that copies already downloaded by third parties prior to a takedown request cannot be recalled, as is intrinsic to any openly licensed public dataset.

\paragraph{Limitations of this ethics regime.}
The participant pool consists of approximately ten human collectors at a single institution and therefore cannot represent the full diversity of body silhouettes, clothing styles, and motion patterns a deployed robot would encounter. This is a substantive limitation of the dataset itself, and is discussed separately in Appendix~\ref{app:ood}.

\newpage
\section*{NeurIPS Paper Checklist}

\begin{enumerate}

\item {\bf Claims}
    \item[] Question: Do the main claims made in the abstract and introduction accurately reflect the paper's contributions and scope?
    \item[] Answer: \answerYes{} 
    \item[] Justification: The abstract and introduction state our main contributions (the HABIT dataset and empirical evidence that training on it elicits human-aware behaviors absent from human-absent demonstration data), and these match the results in Section~\ref{sec:experiments} and the OOD and mid-training analyses in Appendix~\ref{app:ood} and Appendix~\ref{app:mid-training}.

\item {\bf Limitations}
    \item[] Question: Does the paper discuss the limitations of the work performed by the authors?
    \item[] Answer: \answerYes{} 
    \item[] Justification: Limitations are discussed in a dedicated Section~\ref{sec:limitations}, including the single environment, the limited number of human collectors, and the one-to-one human-robot configuration. Additional task-level OOD limitations are discussed in Appendix~\ref{app:ood}, and dataset-level ethical limitations in Appendix~\ref{app:ethics}.

\item {\bf Theory assumptions and proofs}
    \item[] Question: For each theoretical result, does the paper provide the full set of assumptions and a complete (and correct) proof?
    \item[] Answer: \answerNA{} 
    \item[] Justification: The paper does not include theoretical results. The contribution is a dataset and empirical evaluation.

    \item {\bf Experimental result reproducibility}
    \item[] Question: Does the paper fully disclose all the information needed to reproduce the main experimental results of the paper to the extent that it affects the main claims and/or conclusions of the paper (regardless of whether the code and data are provided or not)?
    \item[] Answer: \answerYes{} 
    \item[] Justification: The dataset is released publicly, all per-task setups, evaluation criteria, and protocols are specified in Appendix~\ref{app:task-details}, and the full fine-tuning configurations are tabulated in Appendix~\ref{app:model-training} (Tables~\ref{app:pi05-training} and~\ref{app:groot-training}). The teleoperation pipeline is based on the publicly available DROID codebase~\cite{droid}, and the evaluated VLA models (\(\pi_{0.5}\) and GR00T N1.6) are open-source releases.

\item {\bf Open access to data and code}
    \item[] Question: Does the paper provide open access to the data and code, with sufficient instructions to faithfully reproduce the main experimental results, as described in supplemental material?
    \item[] Answer: \answerYes{} 
    \item[] Justification: The HABIT dataset is publicly released under CC BY 4.0, accompanied by a Croissant metadata file. The teleoperation code derives from the public DROID codebase, and the evaluated VLA models are publicly released open-source checkpoints (\texttt{pi05\_base} and \texttt{nvidia/GR00T-N1.6-3B}). We further provide model finetuning code for HABIT.

\item {\bf Experimental setting/details}
    \item[] Question: Does the paper specify all the training and test details (e.g., data splits, hyperparameters, how they were chosen, type of optimizer) necessary to understand the results?
    \item[] Answer: \answerYes{} 
    \item[] Justification: Training and test details are specified in Section~\ref{sec:exp-main} (training steps, batch size, evaluation cell structure) and in Appendix~\ref{app:model-training} (full hyperparameter tables) and Appendix~\ref{app:task-details} (per-task evaluation protocols).


\item {\bf Experiment statistical significance}
    \item[] Question: Does the paper report error bars suitably and correctly defined or other appropriate information about the statistical significance of the experiments?
    \item[] Answer: \answerNo{} 
    \item[] Justification: We do not report error bars or confidence intervals. Instead, each (task, model, condition) cell is evaluated over $N=20$ trials following a predefined task-specific evaluation protocol (Appendix~\ref{app:eval-protocol}), with the protocol designed to reduce trial-level variance by systematically covering representative task configurations rather than relying on random sampling. This protocol follows standard practice in robot manipulation evaluation, where the substantial physical resources required for additional rollouts make formal variance estimation impractical.

\item {\bf Experiments compute resources}
    \item[] Question: For each experiment, does the paper provide sufficient information on the computer resources (type of compute workers, memory, time of execution) needed to reproduce the experiments?
    \item[] Answer: \answerYes{}  
    \item[] Justification: All training and mid-training runs were performed on a single H100 GPU node, as noted in Appendix~\ref{app:model-training}. Training hyperparameters (training steps, batch size, precision, optimizer settings) are tabulated in Tables~\ref{app:pi05-training} and~\ref{app:groot-training}.
    
\item {\bf Code of ethics}
    \item[] Question: Does the research conducted in the paper conform, in every respect, with the NeurIPS Code of Ethics \url{https://neurips.cc/public/EthicsGuidelines}?
    \item[] Answer: \answerYes{} 
    \item[] Justification: The research conforms with the NeurIPS Code of Ethics. Detailed consent procedures, internal ethics review, risk identification and mitigation, privacy protections, and participant rights are documented in Appendix~\ref{app:ethics}.

\item {\bf Broader impacts}
    \item[] Question: Does the paper discuss both potential positive societal impacts and negative societal impacts of the work performed?
    \item[] Answer: \answerYes{} 
    \item[] Justification: Positive societal impacts (enabling robots that exhibit human-aware behaviors required for safe deployment in shared workspaces) are motivated throughout Section~\ref{sec:intro} and Section~\ref{sec:conclusion}. Negative impacts and limitations (privacy, dataset coverage, OOD failure modes) and the corresponding mitigations are discussed in Section~\ref{sec:limitations} and Appendix~\ref{app:ethics}.
    
\item {\bf Safeguards}
    \item[] Question: Does the paper describe safeguards that have been put in place for responsible release of data or models that have a high risk for misuse (e.g., pre-trained language models, image generators, or scraped datasets)?
    \item[] Answer: \answerYes{} 
    \item[] Justification: All human faces in the public release are blurred, direct identifiers (names, contact information, signed consent forms) are stored separately under restricted access, and a participant takedown procedure with explicit timelines is established. Full data security and access-control measures are documented in Appendix~\ref{app:ethics}.

\item {\bf Licenses for existing assets}
    \item[] Question: Are the creators or original owners of assets (e.g., code, data, models), used in the paper, properly credited and are the license and terms of use explicitly mentioned and properly respected?
    \item[] Answer: \answerYes{} 
    \item[] Justification: All third-party assets are cited with their original publications: DROID~\cite{droid} for the teleoperation codebase, $\pi_{0.5}$~\cite{pi05} and GR00T N1.6~\cite{groot} for the evaluated VLA models. The released dataset uses CC BY 4.0, which is compatible with the open-source licenses of these assets.

\item {\bf New assets}
    \item[] Question: Are new assets introduced in the paper well documented and is the documentation provided alongside the assets?
    \item[] Answer: \answerYes{} 
    \item[] Justification: The HABIT dataset is the new asset introduced in this paper, released under CC BY 4.0 and accompanied by a Croissant metadata file documenting limitations, biases, sensitive information categories, validated use cases, social impact, and provenance activities. Consent procedures from the participants whose data appears in the dataset are described in Appendix~\ref{app:ethics}.

\item {\bf Crowdsourcing and research with human subjects}
    \item[] Question: For crowdsourcing experiments and research with human subjects, does the paper include the full text of instructions given to participants and screenshots, if applicable, as well as details about compensation (if any)? 
    \item[] Answer: \answerYes{} 
    \item[] Justification: Human collectors were full-time staff at the authors' institution, not external crowdworkers. Compensation took the form of regular salary exceeding the local statutory minimum wage, consistent with the NeurIPS Code of Ethics. Participant instructions, training, consent procedures, and right-to-withdraw infrastructure are documented in Appendix~\ref{app:ethics}, and data collection protocols are detailed in Section~\ref{sec:data-collection}.

\item {\bf Institutional review board (IRB) approvals or equivalent for research with human subjects}
    \item[] Question: Does the paper describe potential risks incurred by study participants, whether such risks were disclosed to the subjects, and whether Institutional Review Board (IRB) approvals (or an equivalent approval/review based on the requirements of your country or institution) were obtained?
    \item[] Answer: \answerYes{} 
    \item[] Justification: Prior to data collection, we conducted a documented internal ethics review constituting an equivalent approval based on the requirements of our institution, as explicitly permitted by the NeurIPS Code of Ethics. The review involved three reviewers in distinct roles (principal investigator, independent ethics reviewer outside the data collection team, and data security reviewer) and produced a signed record covering six risk categories. Risks, disclosures to participants, and mitigations are detailed in Appendix~\ref{app:ethics}.

\item {\bf Declaration of LLM usage}
    \item[] Question: Does the paper describe the usage of LLMs if it is an important, original, or non-standard component of the core methods in this research? Note that if the LLM is used only for writing, editing, or formatting purposes and does \emph{not} impact the core methodology, scientific rigor, or originality of the research, declaration is not required.
    \item[] Answer: \answerNA{} 
    \item[] Justification: LLMs are not used as an important, original, or non-standard component of the core methodology. The contribution is a dataset and empirical evaluation of publicly released VLA models.

\end{enumerate}

\end{document}